\documentclass[final,5p,times,twocolumn]{elsarticle}

\usepackage{color}     
\usepackage{flushend}  
\usepackage{graphicx}  
\usepackage{subfigure} 
\usepackage{multirow}  
\usepackage{amsmath}   
\usepackage{amssymb}   
\usepackage{hyperref}  
\usepackage[bottom]{footmisc} 
\usepackage{pbox}      
\usepackage{algorithm2e} 
\usepackage{multirow}

\providecommand{\SetAlgoLined}{\SetLine}
\providecommand{\DontPrintSemicolon}{\dontprintsemicolon}

\hypersetup{
  colorlinks=false,
  pdfborder={0 0 0},
  pdftitle={Real-time Monocular Object SLAM},
  pdfauthor={Dorian G\'alvez-L\'opez, Marta Salas, Juan D. Tard\'os, J. M. M. Montiel},
  pdfkeywords={object slam, object recognition}
}

\DeclareGraphicsExtensions{.png,.jpg,.pdf}  
\graphicspath{{./figures/}}                 

\newcommand{\IGNORE}[1]{}

\newcommand{\etal}{
  \textit{et al}.
}
\DeclareMathOperator*{\argmin}{arg\,min}

\newcommand{\T}{T}                
\newcommand{\Ts}{T^*}  
\newcommand{\R}{R}       
\newcommand{\SSS}{\mathcal{S}}       
\newcommand{\BB}{\mathcal{B}}    
\newcommand{\OO}{\mathcal{O}}    
\newcommand{\tr}{\mathbf{t}}     
\newcommand{\PP}{\mathcal{P}}    
\newcommand{\XX}{\mathcal{X}}    
\newcommand{\xx}{\mathbf{x}}     
\newcommand{\UU}{\mathcal{U}}     
\newcommand{\uu}{\mathbf{u}}     
\newcommand{\kl}{\text{KL}}      
\newcommand{\vv}{\mathbf{v}}     
\newcommand{\ww}{\mathbf{w}}     
\newcommand{\aaa}{\mathbf{a}}     
\newcommand{\ee}{\mathbf{e}}     
\newcommand{\pair}[2]{\langle #1, \, #2 \rangle} 
\newcommand{\triple}[3]{\langle #1, \, #2, \, #3 \rangle} 
\newcommand{\quadruple}[4]{\langle #1, \, #2, \, #3, \, #4 \rangle} 

\newcommand{\figQueryScores}
{
  \begin{figure}[!t]
  \centering
  \subfigure[]{
    \includegraphics[width=0.47\columnwidth]{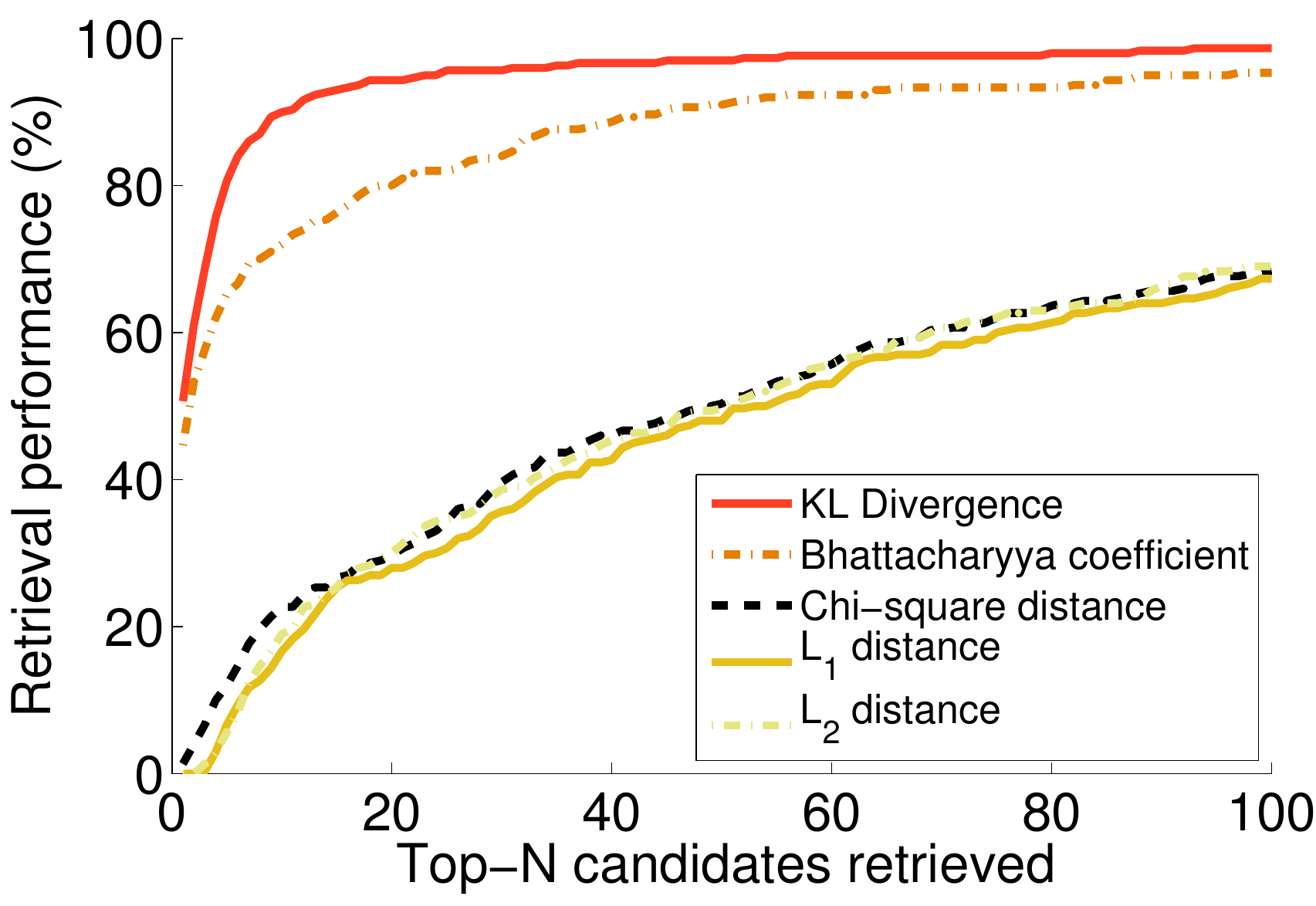}
    \label{fig:QueryScores:Acc}
  }
  \subfigure[]{
    \includegraphics[width=0.47\columnwidth]{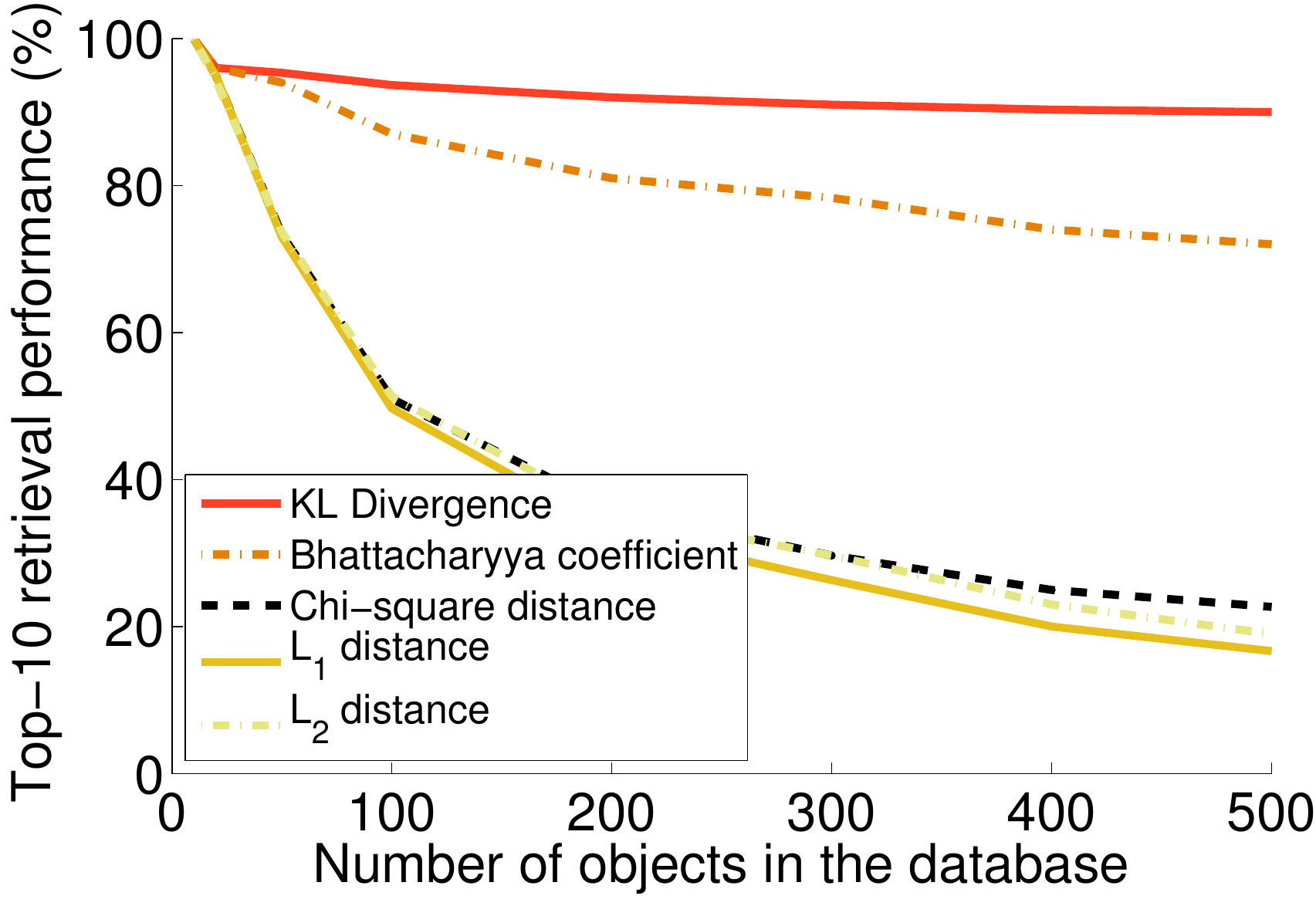}
    \label{fig:QueryScores:Nobjs}
  }
  \caption{Performance of several similarity scores to retrieve candidates 
    when
    \subref{fig:QueryScores:Acc} a database with 500 objects is queried, and
    \subref{fig:QueryScores:Nobjs} the top-10 candidates are retrieved
    from databases of different sizes.}
  \label{fig:QueryScores}
  \end{figure}
}

\newcommand{\figClusteringQuery}
{
  \begin{figure*}[!t]
    \subfigure[Putative correspondences from the entire image]
    {
      \includegraphics[width=0.5\textwidth]{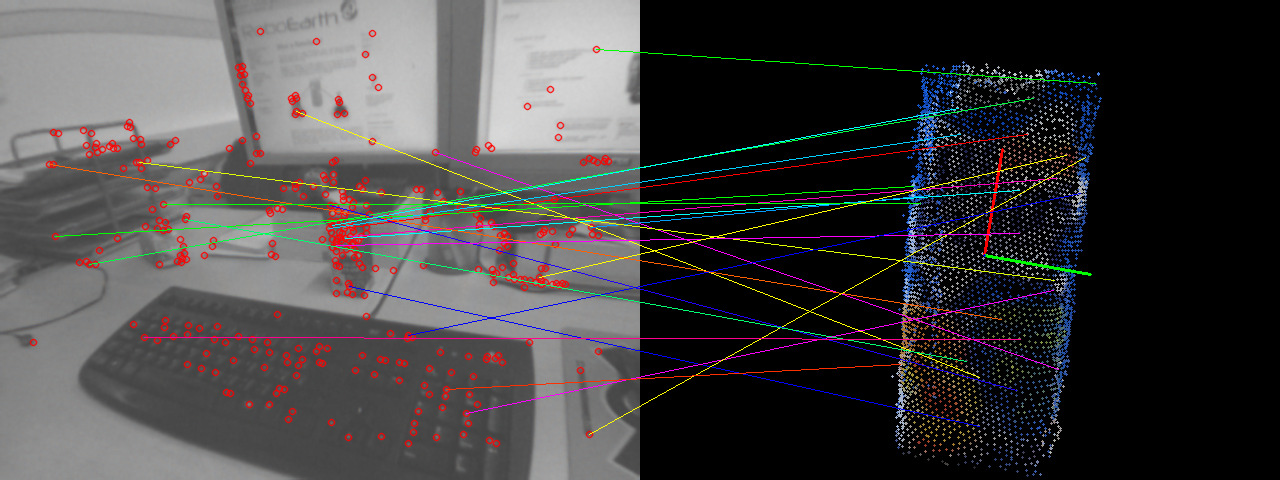}
      \label{fig:ClusteringQuery-noclusters}
    }
    \subfigure[Putative correspondences from regions]
    {
      \includegraphics[width=0.5\textwidth]{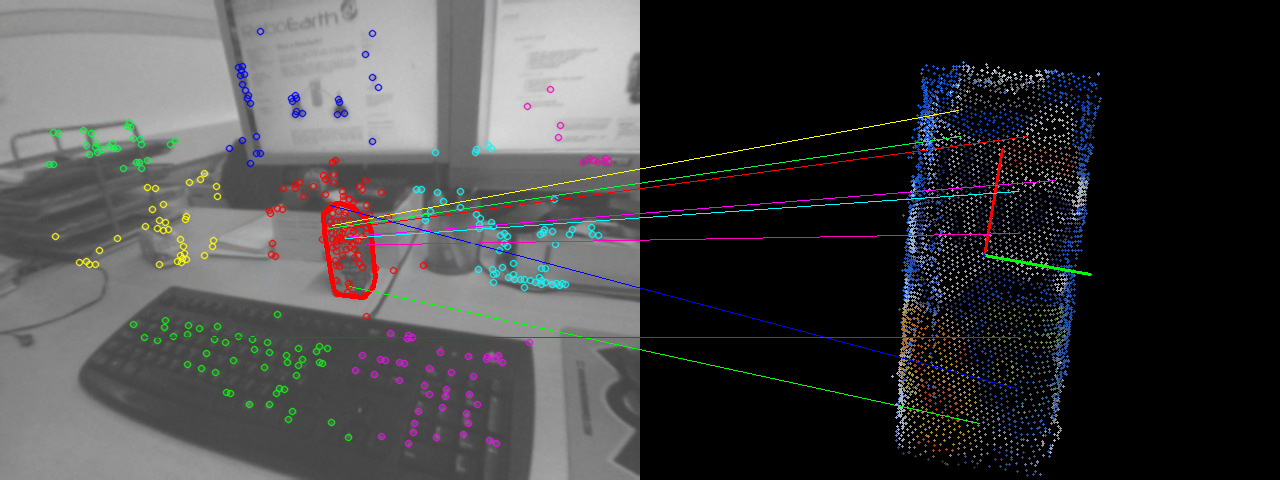}
      \label{fig:ClusteringQuery-clusters}
    }
    
    \caption{Example of putative correspondences obtained in an image with 356 
    features. 
    In \subref{fig:ClusteringQuery-noclusters}, all the features are
    used to query an object database of 500 models and to compute correspondences. 
    The correct object is the 7th candidate after querying, and 24 putative 
    correspondences are computed, where 18 are incorrect. The object pose
    cannot be successfully verified after 100 random iterations trying
    sets of correspondences. 
    On the other hand, in \subref{fig:ClusteringQuery-clusters}, regions of
    features are used individually to query the database and to produce
    putative correspondences. In this case, the correct object appears in the
    1st position (out of 500) when the region that contains it is queried. In total, 9
    correspondences are computed, where just 3 are incorrect. This makes it
    possible to verify the object and obtain its pose after 34 iterations. 
    }
    \label{fig:ClusteringQuery}
  \end{figure*}
}

\newcommand{\figClusteringVan}
{
  \begin{figure}[t]
    \includegraphics[width=\columnwidth]{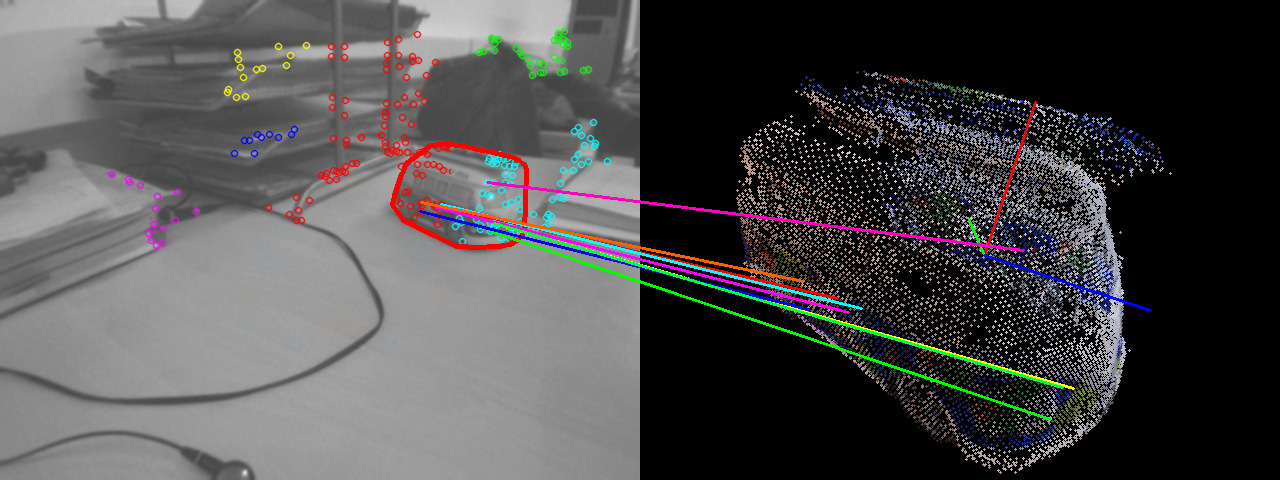}
    \caption{Example of a real object that lies in two different image regions.
    Since putative correspondences from different regions are merged,
    it is correctly found with 9 inliers. } 
    \label{fig:ClusteringVan}
  \end{figure}
} 

\newcommand{\figRansacSteps}
{
  \begin{figure}
  \subfigure[Some randomly selected putative correspondences yield a first pose.]
  {
    \includegraphics[width=\columnwidth]{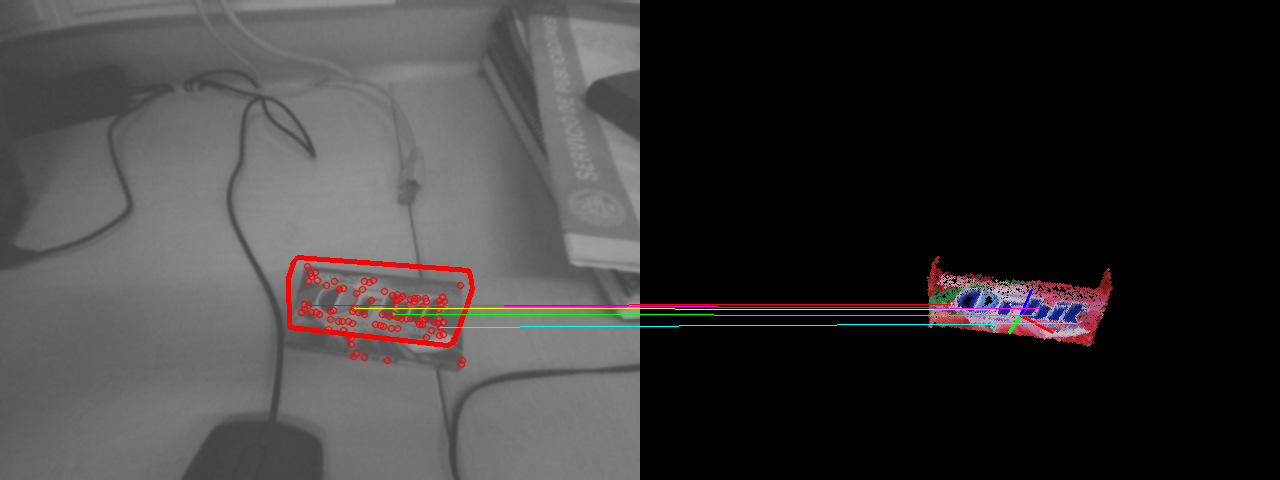}
    \label{fig:RansacSteps-putative}
  }
  \subfigure[A final pose is computed from the new correspondences obtained after
    projecting the object model.]
  {
    \includegraphics[width=\columnwidth]{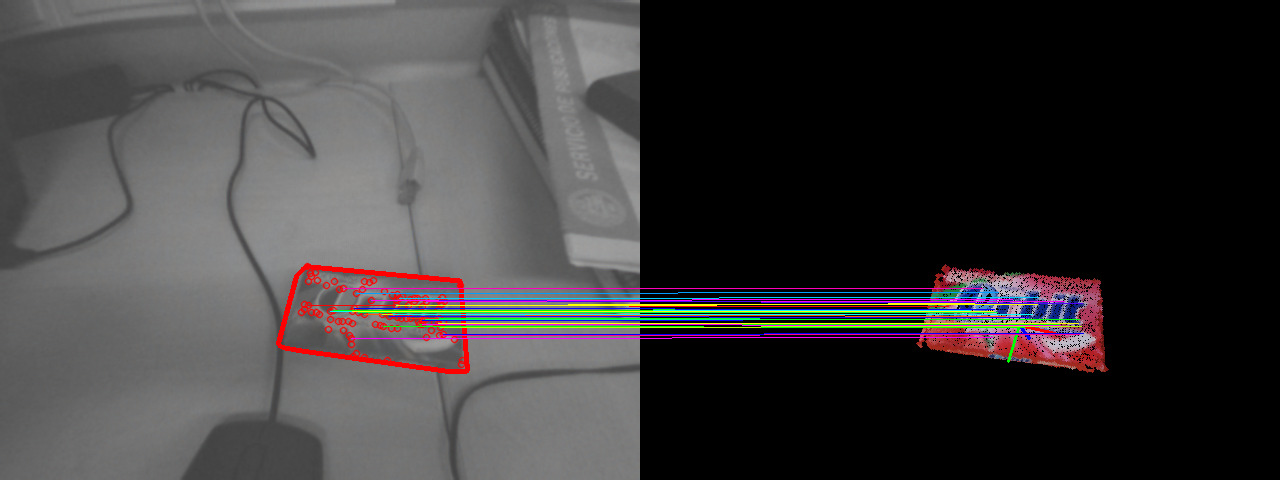}
    \label{fig:RansacSteps-projection}
  }
  \caption{Computation of a robust candidate pose in a DISAC iteration.}
  \label{fig:RansacSteps}
  \end{figure}
}

\newcommand{\figOverview}
{
  \begin{figure*}[t]
  \includegraphics[width=\textwidth]{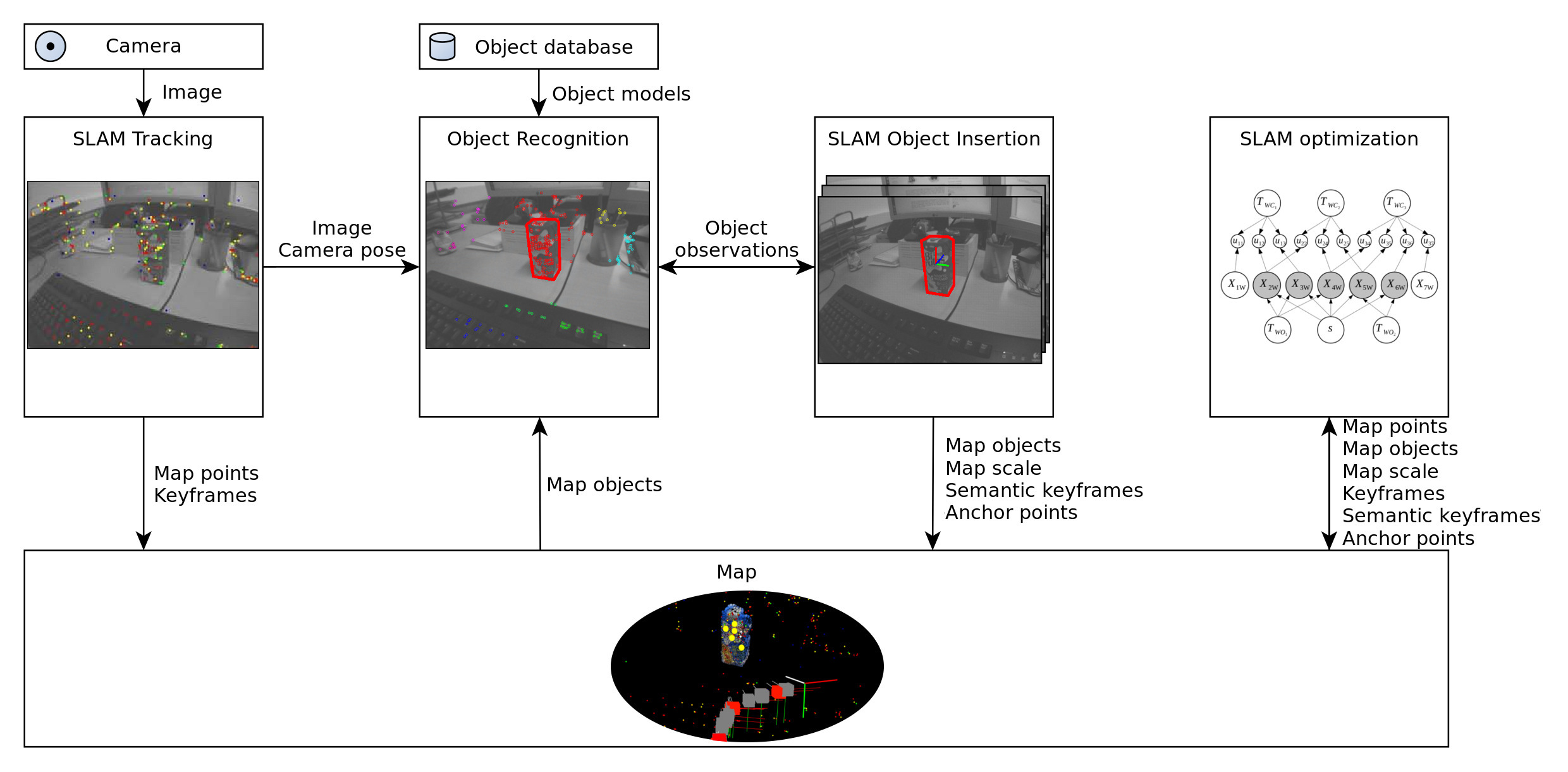}
  \caption{
  System overview:
  Every video frame is processed by the SLAM tracking thread to locate the camera, 
  and to determine if a new keyframe 
  is added to the map.
  Object recognition is applied to as many frames as possible,
  exploiting the information of the location of objects previously seen.
  If the recognition is successful, the observation of the object is stored
  until there is enough geometrical information about it.
  In that moment, the object instance is triangulated and
  inserted in the map, together with new map points anchored to object points and
  a subset of frames that observed them, coined semantic keyframes. 
  This operation allows to find the map scale and to include object geometrical 
  constraints to the map optimization. 
  }
  \label{fig:Overview}
  \end{figure*}
}

\newcommand{\figBayesianNetworks}
{
\begin{figure}
\centering
 \subfigure[Standard SLAM]{
   \includegraphics[width=0.60\columnwidth]{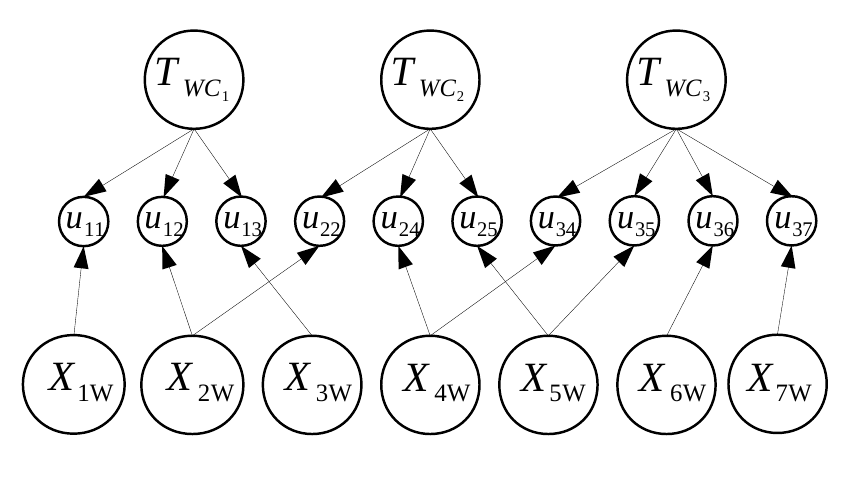}
   \label{fig:BayesianSlam}
 }

 \subfigure[Object SLAM]{
  \includegraphics[width=0.60\columnwidth] {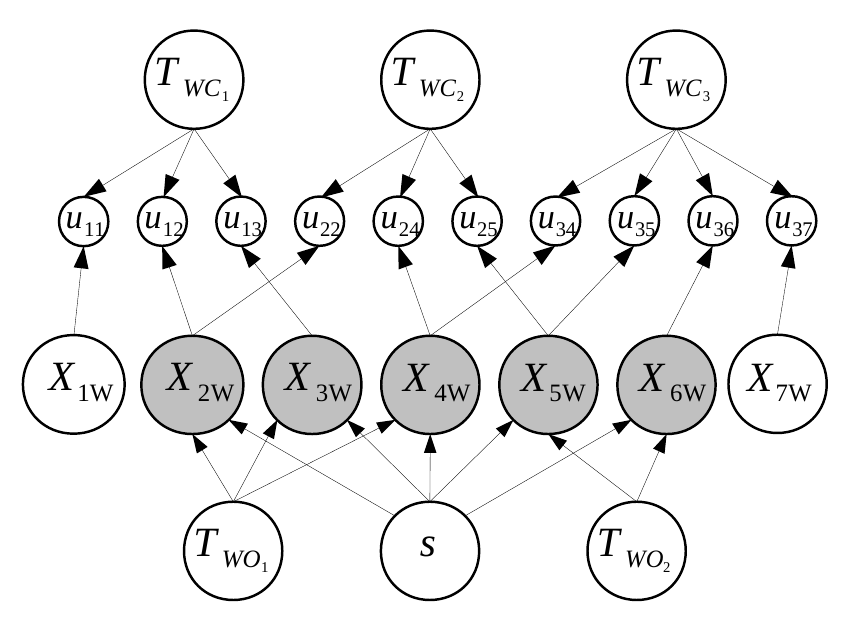}
   \label{fig:BayesianObjectSlam}

}
\caption{ SLAM estimation problem. 
a) Bayesian network of standard SLAM. 
$\T_{WC_i}$ are the cameras, $\xx_{jW}$ the map points and $\uu_{ij}$ is 
the measurement on the image.  
b)  Bayesian network of object SLAM. Some objects are added to the BA, where the 
object location is represented as $\T_{WO_k}$ and the scale $s$ becomes  
observable. Highlighted map points are those which belong to the objects.
}
\label{fig:BayesianNetwork}
\end{figure}
}

\newcommand{\figTriangulation}
{
  \begin{figure}
  \centering
  \includegraphics[width=\columnwidth]{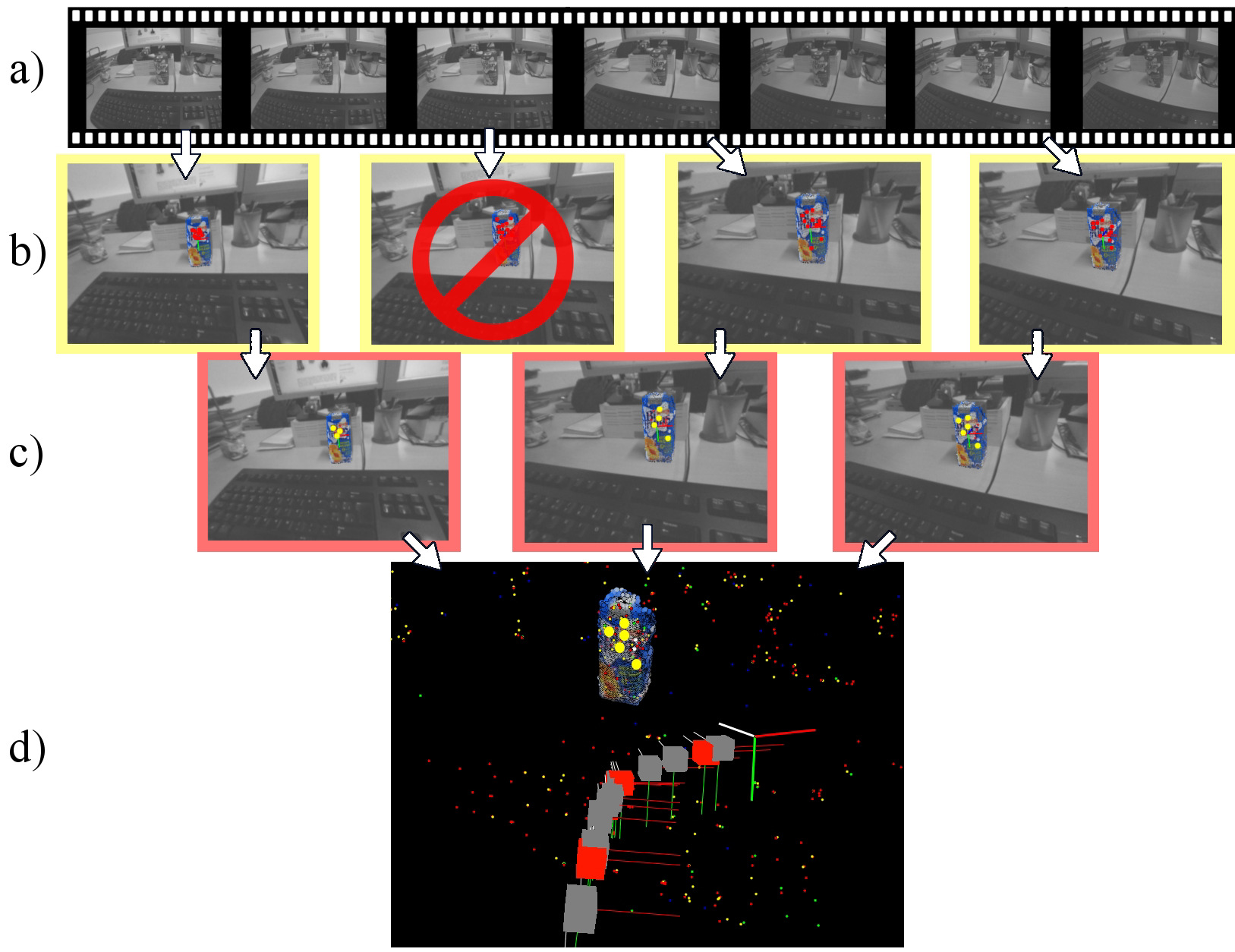}
  \caption{Object insertion with a monocular camera. 
  a) Object detection is performed as fast as possible on the frames of the
  video stream. 
  b) The bottle is detected in some frames and its observed 2D
  points (red points) are accumulated. 
  c) When several points are observed with enough parallax (yellow points),
  their frames are selected as semantic keyframes. Detection frames that offer
  no parallax are discarded. 
  d) Observations from semantic keyframes are used to triangulate the object and
  its 3D points. The semantic keyframes (red cameras), the object and its points 
  are inserted in
  the map, updating its scale. }
  \label{fig:triangulation}
  \end{figure}
}

\newcommand{\figScaleEstimation}
{
\begin{figure}
\centering

  \includegraphics[width=0.75\columnwidth] {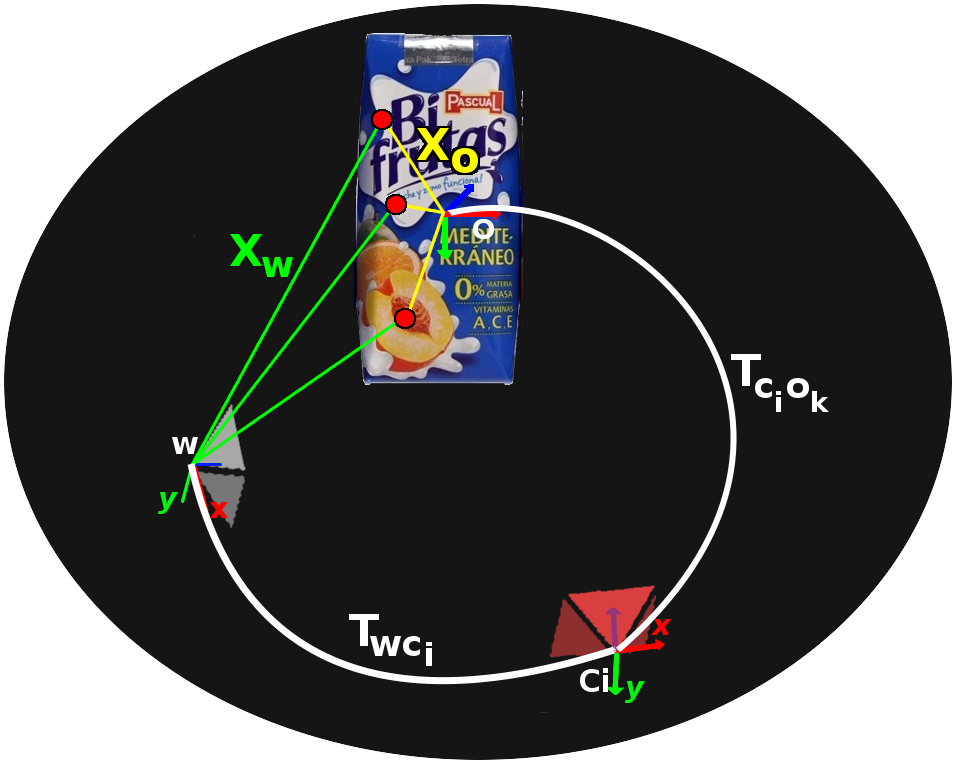}

\caption{ Object pose prior estimation ($T_{WO_k}$). Red landmarks show the pairs $\pair{\xx_O}{\xx_W}$ of object points and anchor points. $T_{WC_i}$ is the pose of the camera, with map scale $s$ already estimated, and $T_{C_iO_k}$ is the pose of the object with respect to the camera for the last observation $B^i_{O_k}$.}
\label{fig:ScaleEstimation}
\end{figure}
}

\newcommand{\figTestingObjects}
{
  \begin{figure}[t]
  \centering
  \includegraphics[width=\columnwidth]{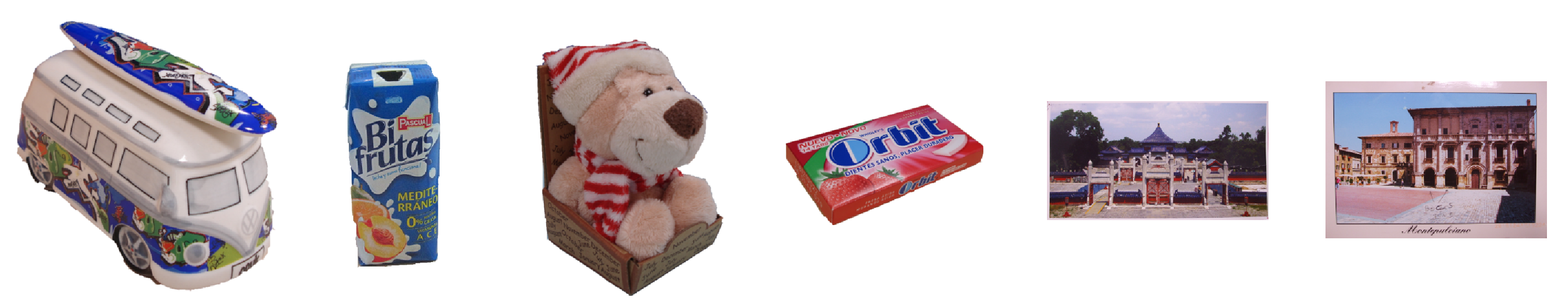}
  \caption{Testing objects of the desktop dataset}
  \label{fig:TestingObjects}
  \end{figure}
}

\newcommand{\figNisterObjects}
{
  \begin{figure}[!t]
  \centering
  \includegraphics[width=0.25\columnwidth]{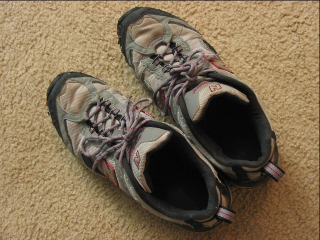}
  \includegraphics[width=0.25\columnwidth]{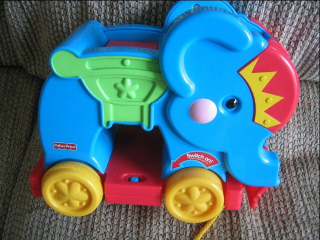}
  \includegraphics[width=0.25\columnwidth]{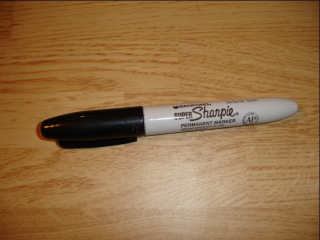}
  \caption{Example of Nister \& Stewenius's \cite{Nister06} objects}
  \label{fig:NisterObjects}
  \end{figure}
}

\newcommand{\figExperimentDesktopDetections}
{
  \begin{figure}[t]
  \centering
  \includegraphics[width=0.235\textwidth]{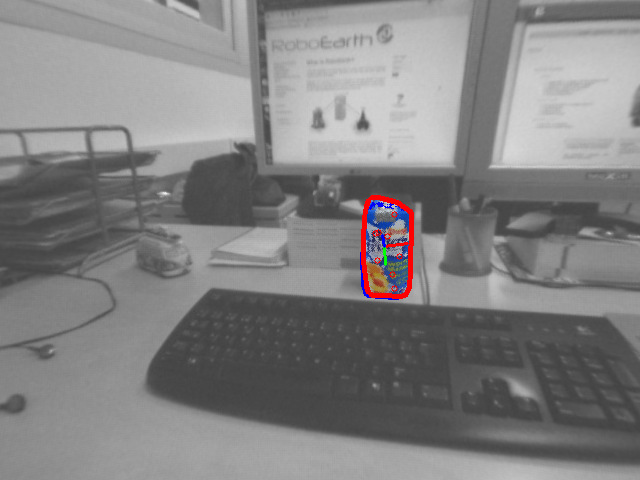}
  \includegraphics[width=0.235\textwidth]{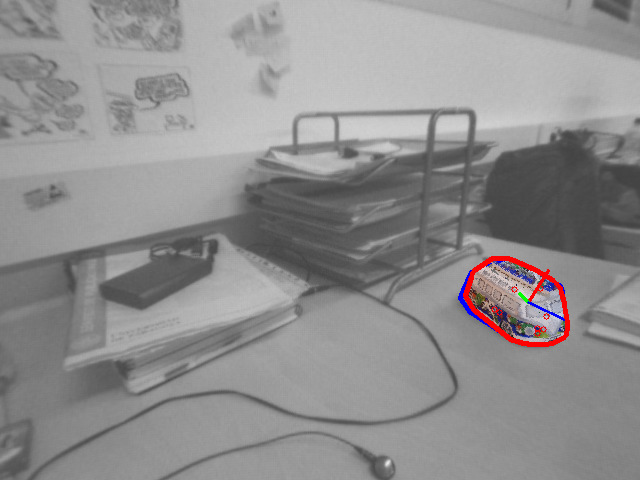}
  \includegraphics[width=0.235\textwidth]{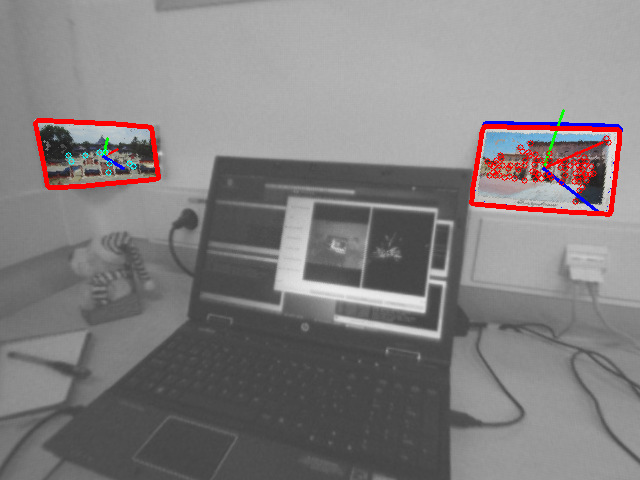}
  \includegraphics[width=0.235\textwidth]{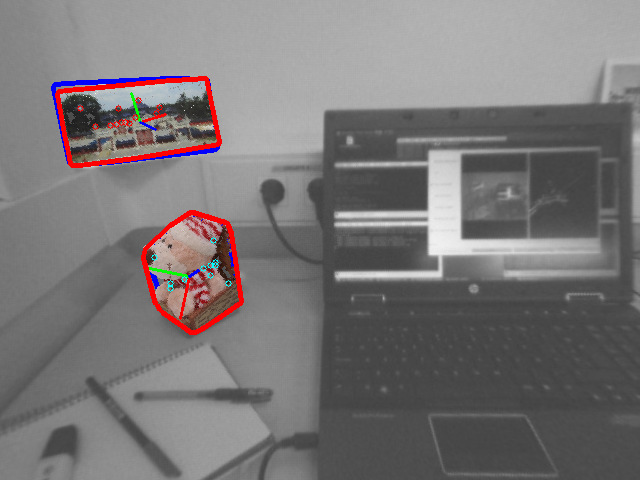}
  \caption{Example of correct detections in the desktop dataset, with
    500 objects in the database.}
  \label{fig:ExperimentDesktopDetections}
  \end{figure}
}

\newcommand{\figExperimentDesktopMap}
{
  \begin{figure}[t]
  \centering
    \includegraphics[width=0.235\textwidth]{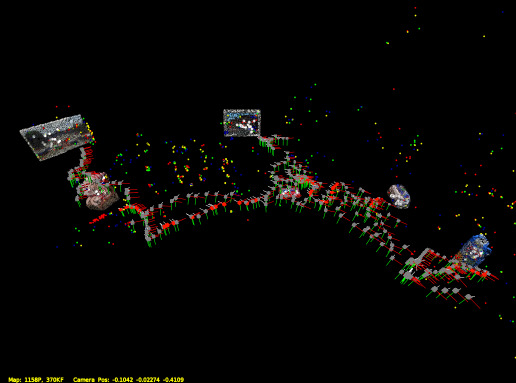}
    \includegraphics[width=0.235\textwidth]{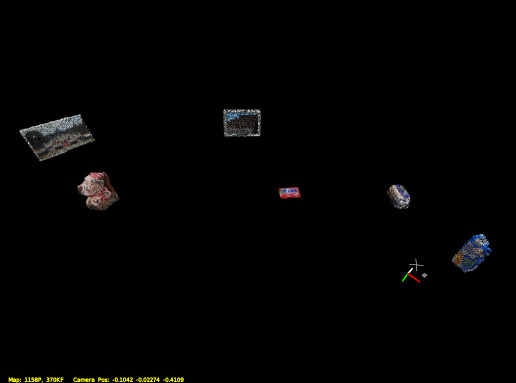}
    \includegraphics[width=0.235\textwidth]{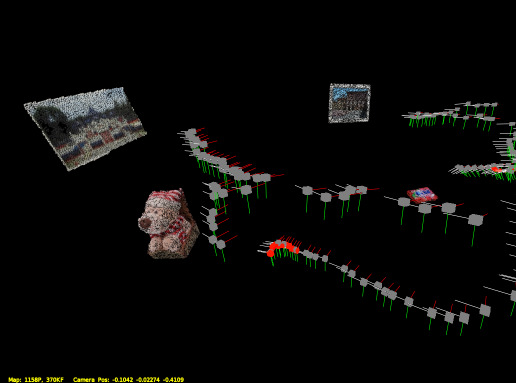}
    \includegraphics[width=0.235\textwidth]{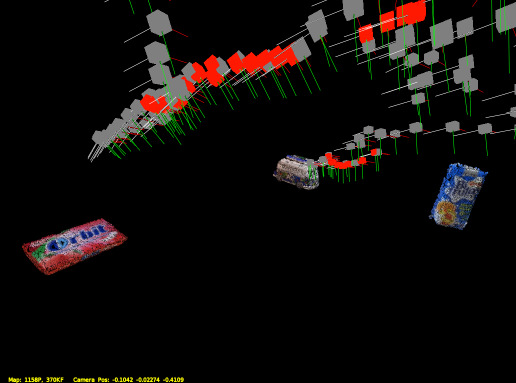}
  \caption{Resulting map of the desktop dataset with 500 models in the database.
  All the objects are correctly located in the space with no false positives.}
  \label{fig:ExperimentDesktopMap}
  \end{figure}
}

\newcommand{\figExecutionTime}
{
  \begin{figure}[t]
  \centering
  \subfigure[SLAM Tracking]{
    \includegraphics[width=.8\columnwidth]{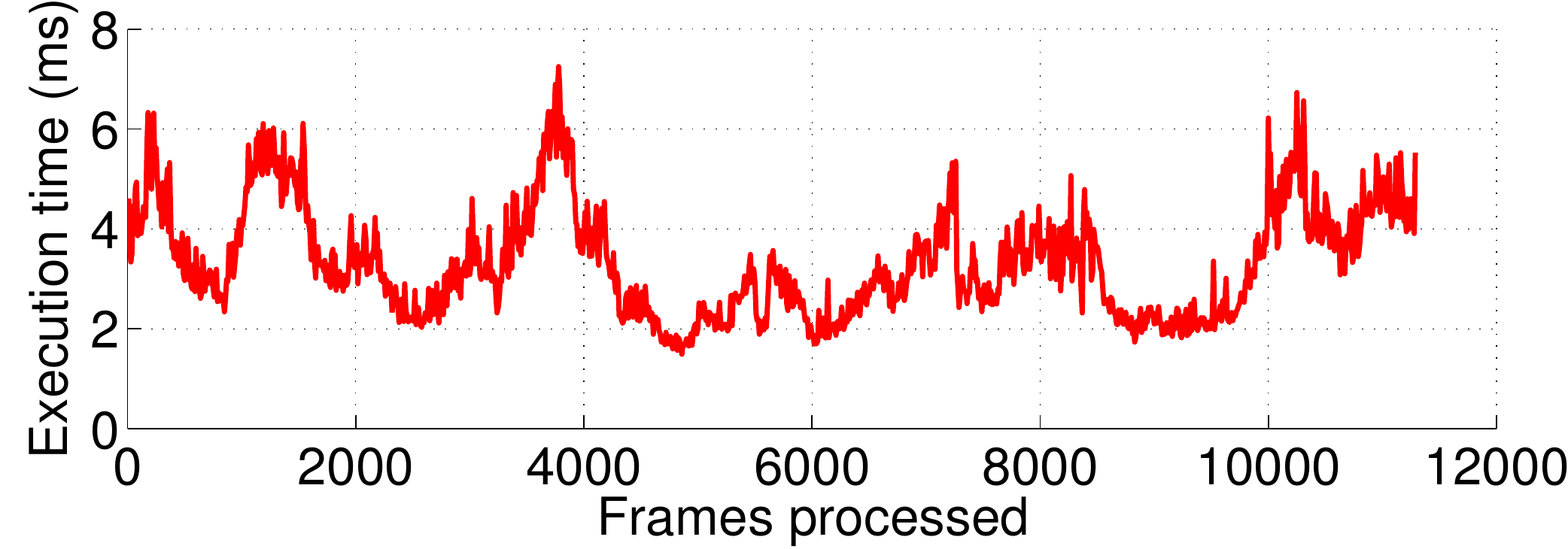}
    \label{fig:ExecutionTimeTracking}}
  \subfigure[Object recognition]{
    \includegraphics[width=.8\columnwidth]{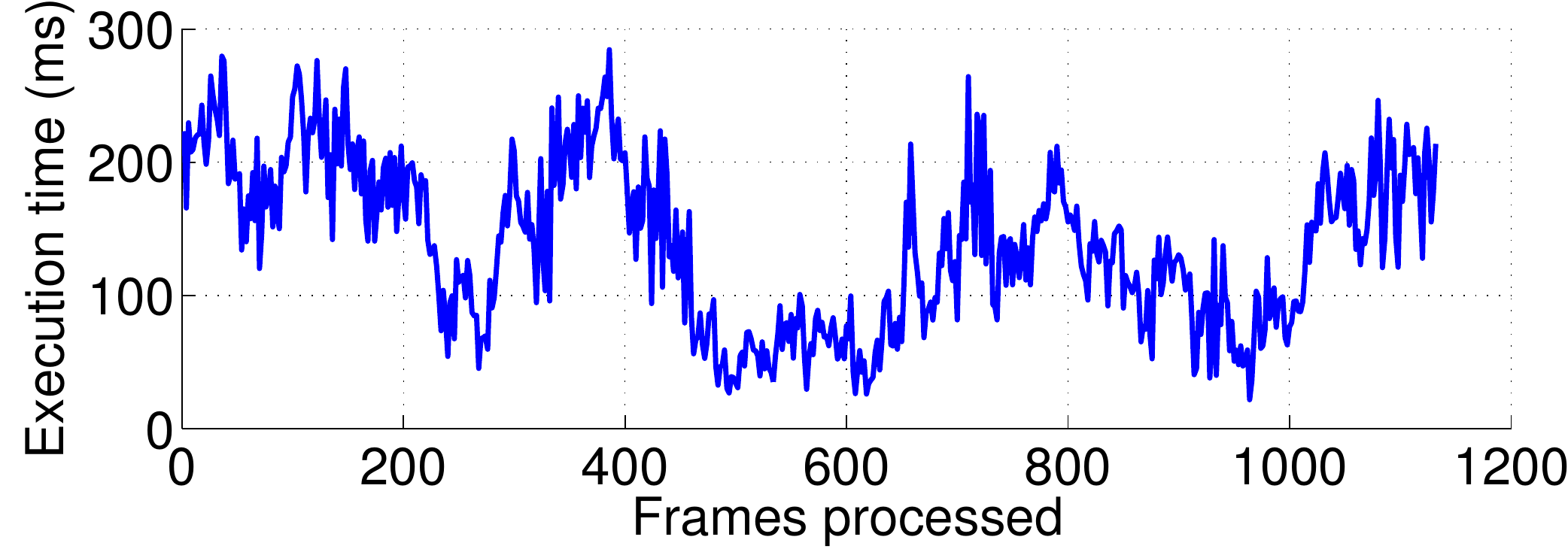}
    \label{fig:ExecutionTimeRecognition}}
  \caption{System execution time in the Desktop dataset}
  \label{fig:ExecutionTime}
  \end{figure}
}

\newcommand{\tabExecutionTime}
{
  \begin{table}[t]
  \centering
  
  \begin{tabular}{lcc}
  & Median & Max. \\
  \hline
  Our system (Object recognition + SLAM) & $0.14$ & $0.34$ \\
  MOPED \cite{Collet2011} (Object recognition) & $0.52$ & $0.95$ \\
  \end{tabular}
  
  \caption{Execution time of the object recognition stage of our system compared
  with MOPED, with a database of 500 objects (s/image). }
  \label{tab:ExecutionTime}
  \end{table}
}

\newcommand{\tabPriorsMoped}
{
  \newcommand{\OurSystemNoPriors}
  {
    \pbox{\textwidth}{\relax\ifvmode\centering\fi 
                    Our system, \\ no priors}
  }
  \begin{table}[t]
  \centering
  \begin{tabular}{lccc}
                  & Our system  & \OurSystemNoPriors & MOPED \cite{Collet2011} \\
  \hline
  Bottle          & $137$       & $41$                  & $81$ \\ 
  Van toy         & $104$       & $33$                  &  $8$ \\ 
  Box             &  $91$       & $56$                  & $63$ \\ 
  Lion toy        &  $80$       & $5$                   & $0$ \\
  Card 1          & $258$       & $227$                 & $166$ \\ 
  Card 2          & $200$       & $118$                 & $121$ \\ 
  \hline
  Total detections & $870$      & $480$                 & $439$ \\ 
  \end{tabular}
  \caption{Number of detections in the desktop dataset
    with a database of 500 objects. By exploiting the priors given by the
    SLAM map we can provide more detections, even compared with
    MOPED, a state-of-the-art single-image recognition algorithm. }
  \label{tab:PriorsMoped}
  \end{table}
}

\newcommand{\figDetectionError}
{
  \begin{figure}[t]
  \centering
  \subfigure[]{
    \includegraphics[width=0.45\columnwidth]{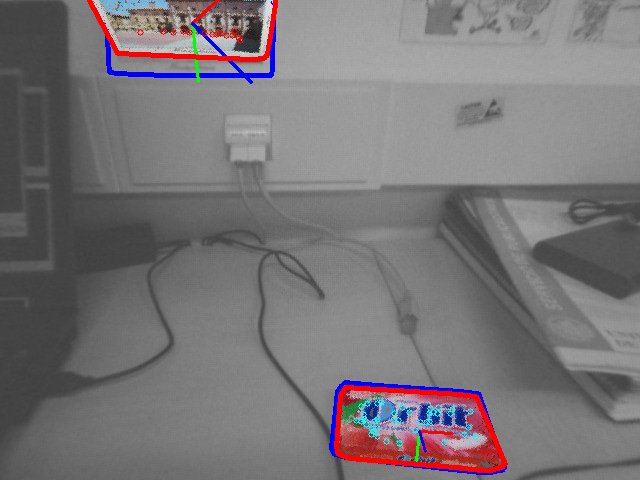}
    \label{fig:DetectionErrorA}}
  \subfigure[]{
    \includegraphics[width=0.45\columnwidth]{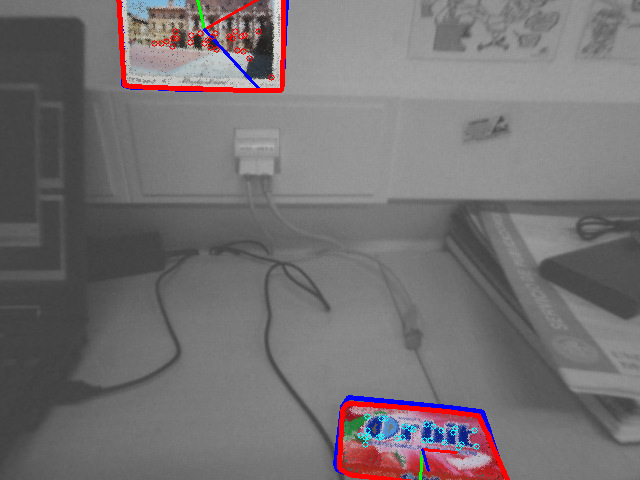}
    \label{fig:DetectionErrorB}}
  \caption{Example of robustness against inaccurate detections. 
  a) Prior knowledge about the location of the objects (blue outline) is used
  to recognize objects. 
  Some features (red dots) of the card are correctly matched with the model, but these are
  ill-distributed and the pose calculated in this single frame is inaccurate.
  b) In the next frame, the actual pose of the card remains correct because it
  is computed from all the accumulated observations. This allows to accurately detect
  the card again.
  }
  \label{fig:DetectionError}
  \end{figure}
}

\newcommand{\figExecutionTimeRgbd}
{
  \begin{figure}[!b]
  \centering
  \subfigure[SLAM Tracking]{
    \includegraphics[width=.8\columnwidth]{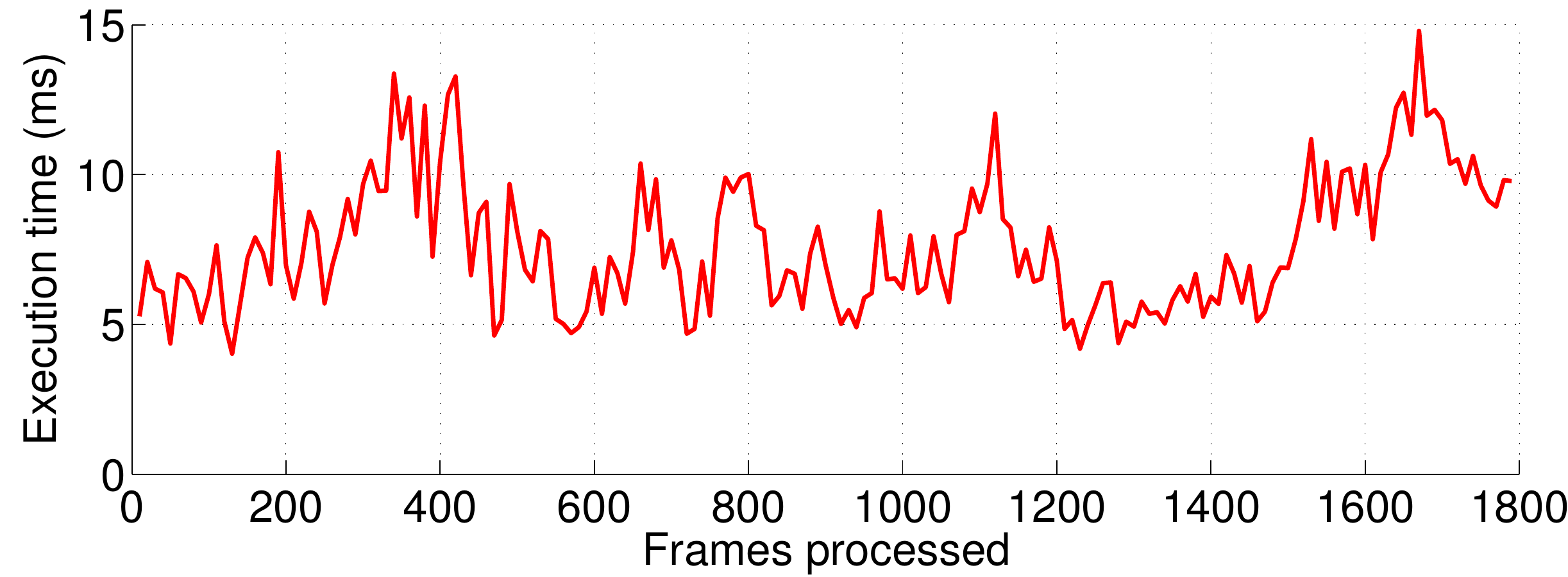}
    \label{fig:ExecutionTimeTrackingRgbd}}
  \subfigure[Object recognition]{
    \includegraphics[width=.8\columnwidth]{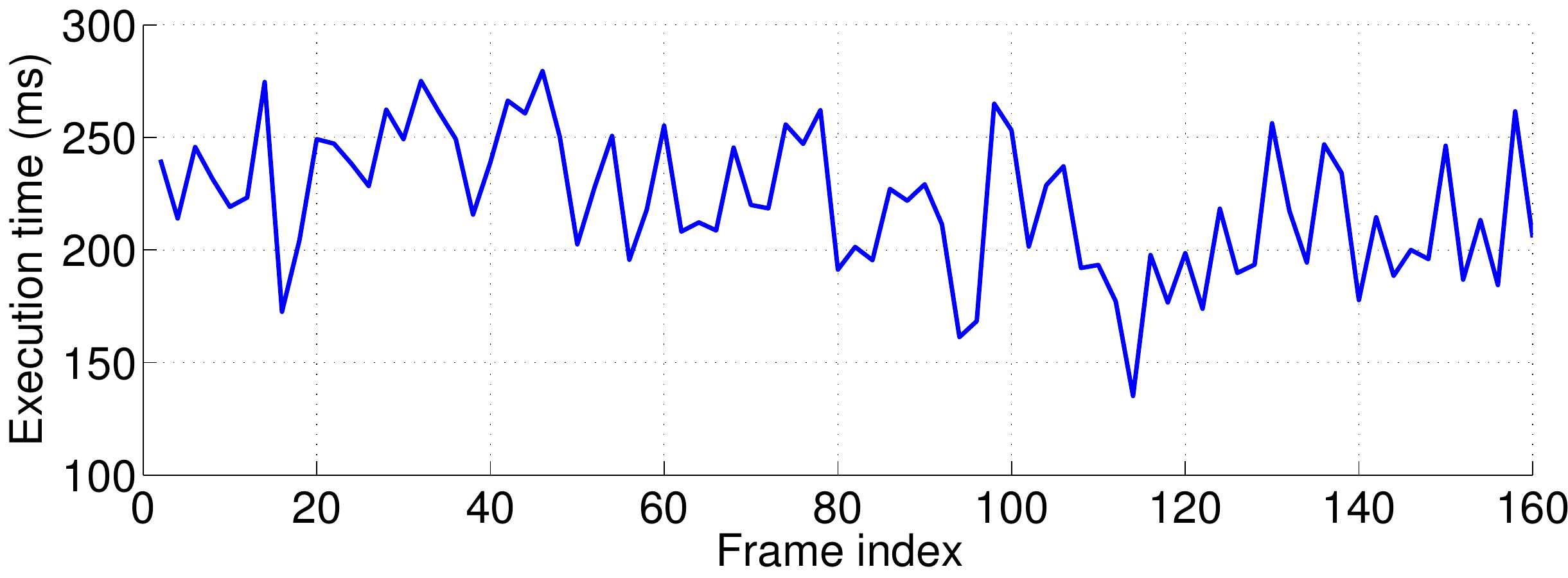}
    \label{fig:ExecutionTimeRecognitionRgbd}}
  \caption{System execution time in the RGB-D SLAM dataset}
  \label{fig:ExecutionTimeRgbd}
  \end{figure}
}

\newcommand{\figAroasRoom}
{
  \begin{figure}[t]
  \centering
  \includegraphics[width=0.75\columnwidth]{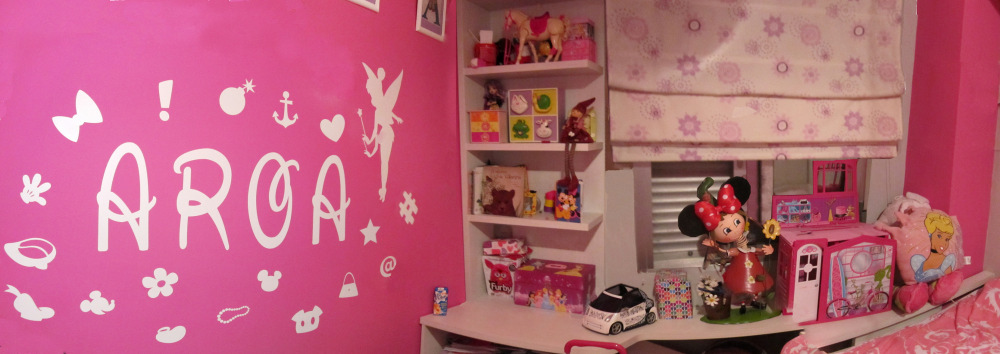}
  \includegraphics[width=0.23\columnwidth]{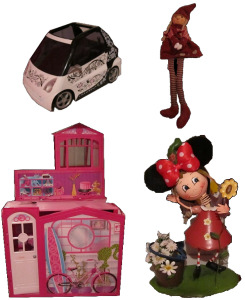}
  \caption{Aroa's room and 4 of the 13 objects modeled}
  \label{fig:AroasRoom}
  \end{figure}
}

\newcommand{\figAroasRoomMap}
{
  \begin{figure}[t]
  \centering
  \includegraphics[width=\columnwidth]{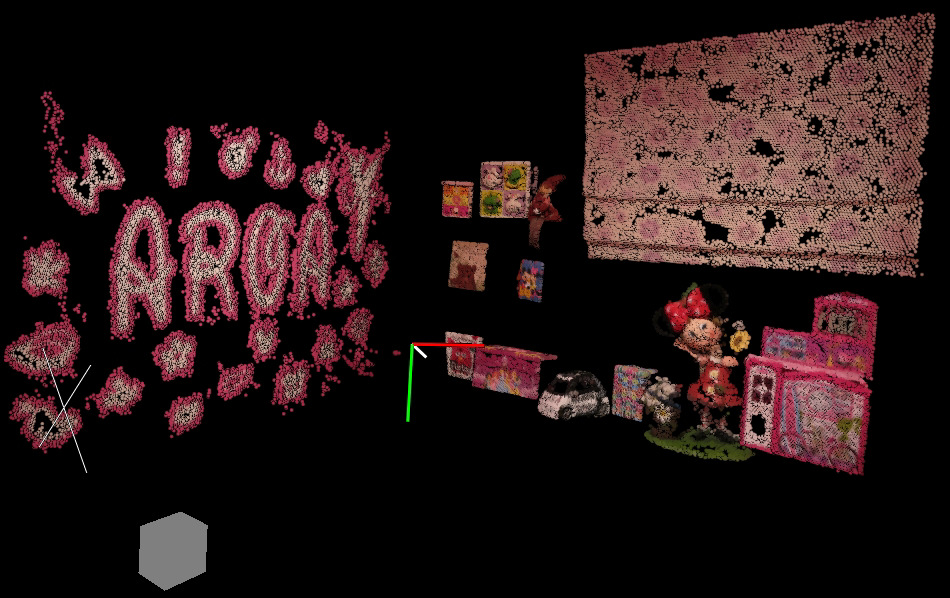}
  \caption{Map of objects created in Aroa's room}
  \label{fig:AroasRoomMap}
  \end{figure}
}

\newcommand{\figSnack}
{
  \begin{figure}[t]
  \centering
  \includegraphics[width=0.8\columnwidth]{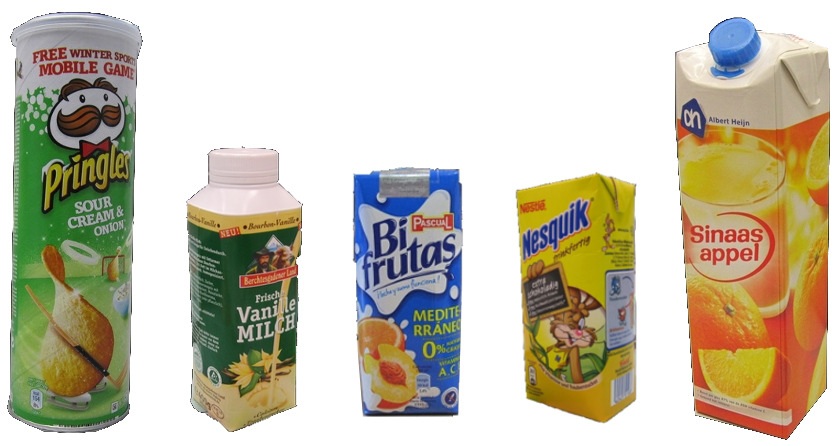}
  \caption{The 5 models out of 21 that appear in the Snack dataset}
  \label{fig:Snack}
  \end{figure}
}

\newcommand{\figSnackPTAM}
{
  \begin{figure}[t]
  \centering
  \includegraphics[width=\columnwidth]{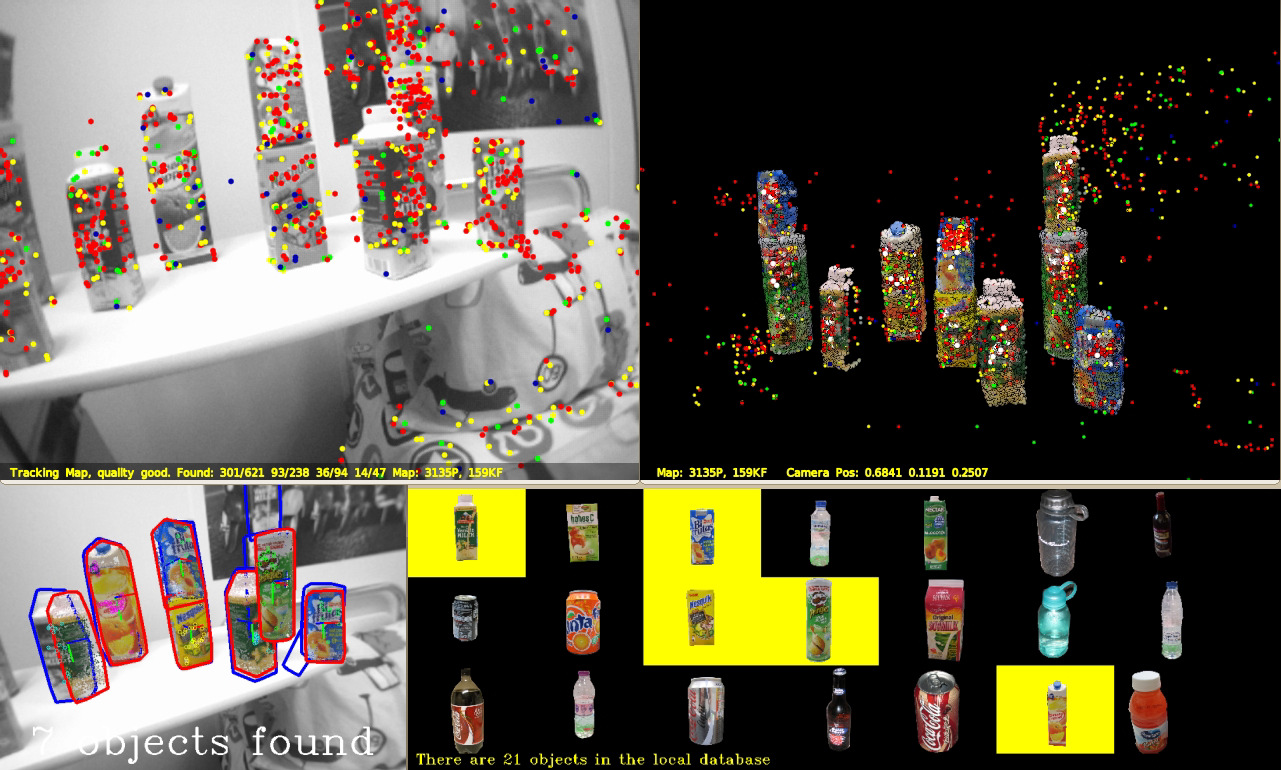}
  \caption{Object SLAM running in the Snack dataset}
  \label{fig:SnackPTAM}
  \end{figure}
}

\newcommand{\figSnackMap}
{
  \begin{figure}[!t]
  \centering
  \includegraphics[width=\columnwidth]{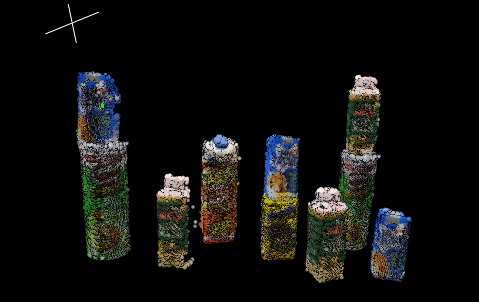}
  \caption{Map of objects created in the Snack dataset}
  \label{fig:SnackMap}
  \end{figure}
}

\newcommand{\figSnackClutter}
{
  \begin{figure}[t]
  \centering
  \includegraphics[width=0.8\columnwidth]{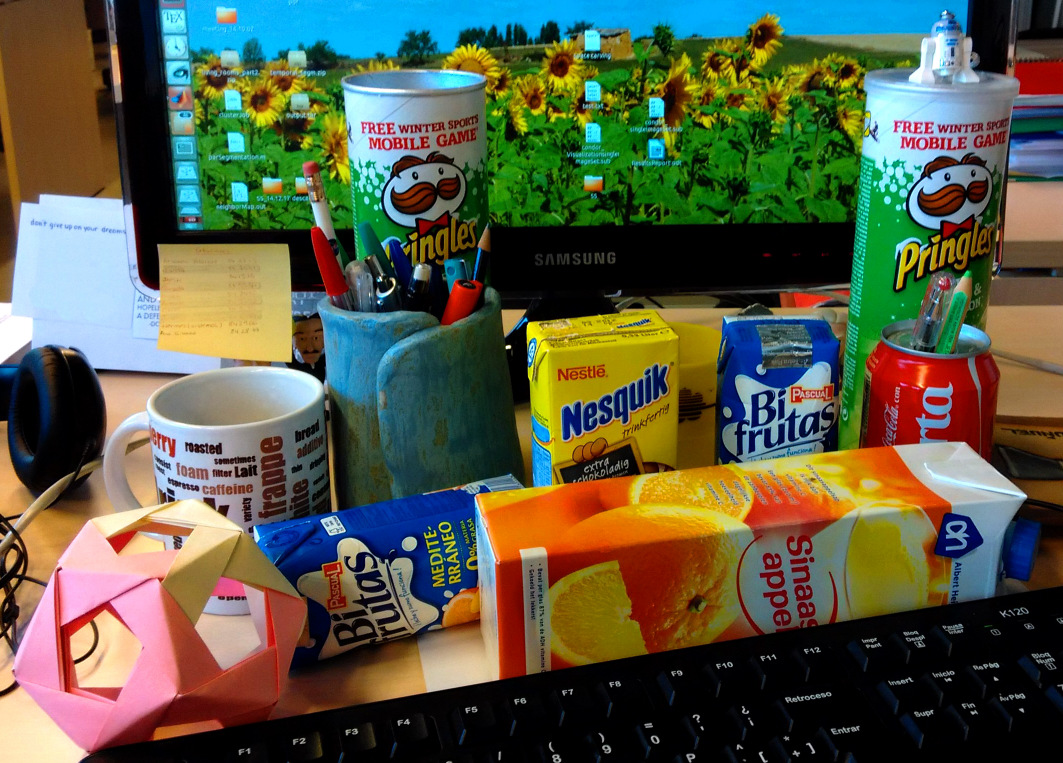}
  \caption{Snack with clutter scenario}
  \label{fig:SnackClutter}
  \end{figure}
}

\newcommand{\figSnackClutterMap}
{
  \begin{figure}[t]
  \centering
  \includegraphics[width=0.8\columnwidth]{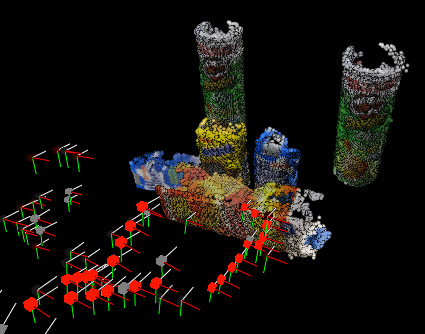}
  \caption{Map of objects created in the Snack with clutter dataset}
  \label{fig:SnackClutterMap}
  \end{figure}
}

\newcommand{\figSnackClutterOk}
{
  \begin{figure}[!t]
  \centering
  \subfigure[An occluded bottle is detected without prior information]{
    \includegraphics[width=\columnwidth]{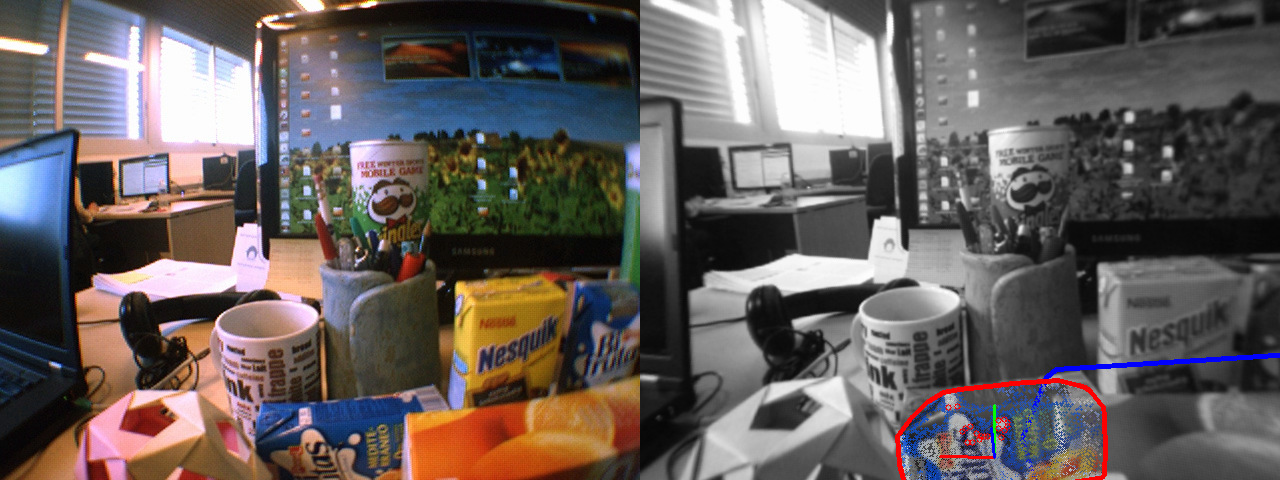}
    \label{fig:SnackClutterOk:Bio}
  }
  \subfigure[An occluded can is detected without prior information]{
    \includegraphics[width=\columnwidth]{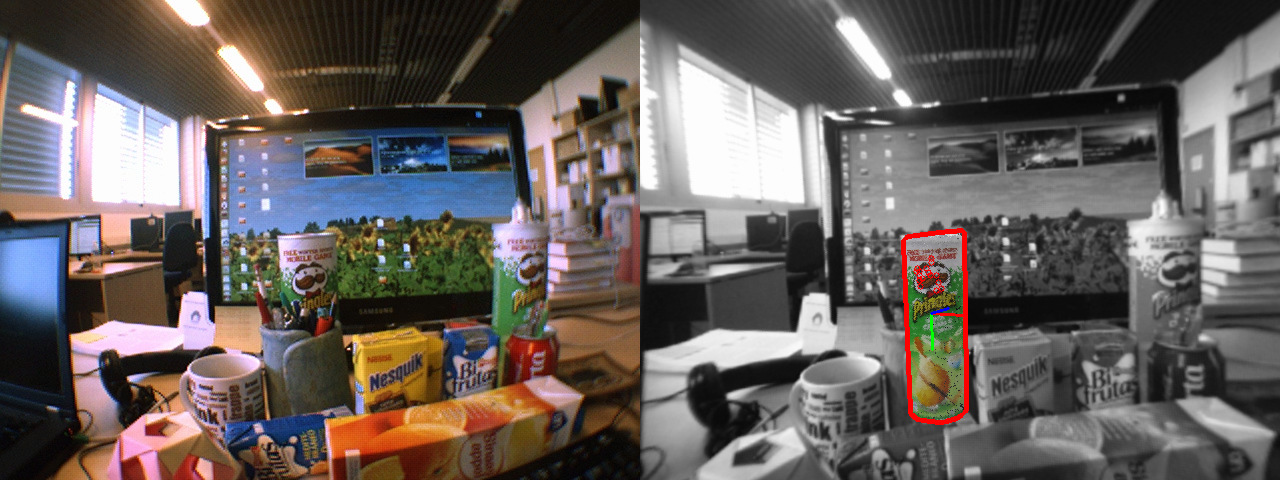}
    \label{fig:SnackClutterOk:Pringles}
  }
  \subfigure[All objects are detected with prior information]{
    \includegraphics[width=\columnwidth]{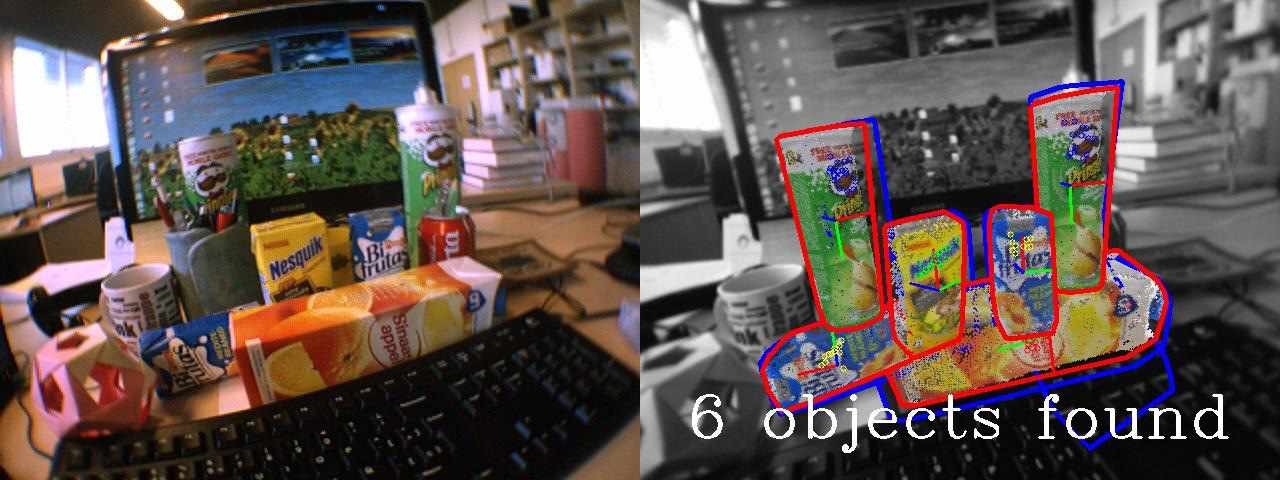}
    \label{fig:SnackClutterOk:All}
  }
  \caption{Successful detections in the Snack with clutter scenario}
  \label{fig:SnackClutterOk}
  \end{figure}
}

\newcommand{\figSnackClutterBad}
{
  \begin{figure}[t]
  \centering
  \includegraphics[width=\columnwidth]{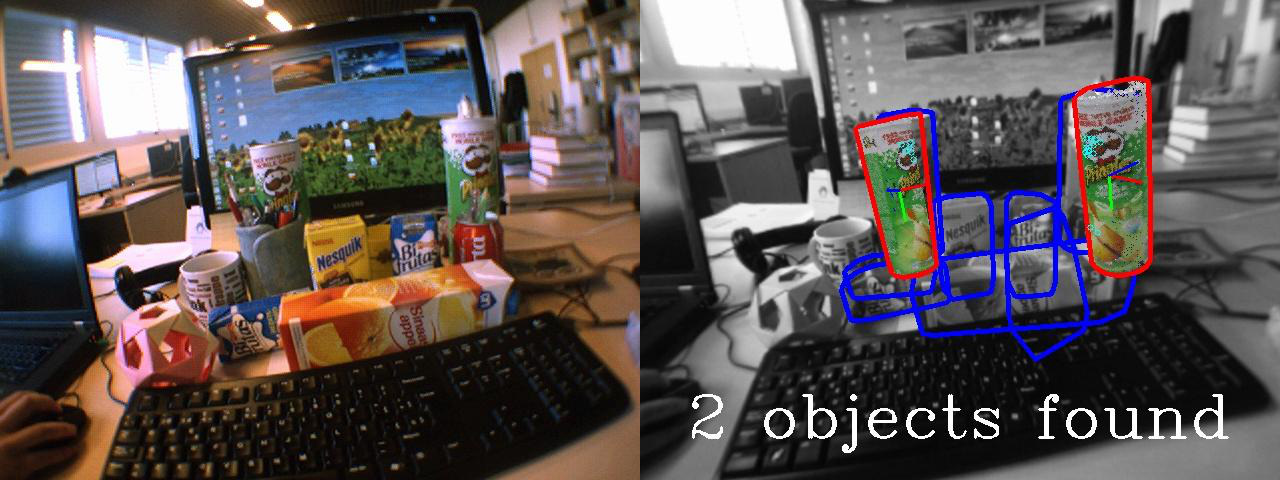}
  \caption{Wrong priors are created around the blue bottles by inaccurate detections,
  but they are are not triangulated and the map remains correct.}
  \label{fig:SnackClutterBad}
  \end{figure}
}

\newcommand{\figRGBDSLAM}
{
  \begin{figure}[!b]
  \centering
  \includegraphics[width=0.8\columnwidth]{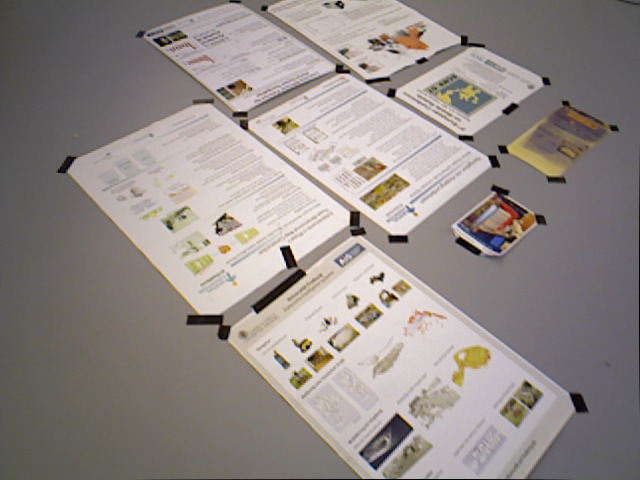}
  \caption{Objects of the RGB-D SLAM Dataset}
  \label{fig:RGBDSLAM}
  \end{figure}
}

\newcommand{\tabGTerror}
{
  \begin{table}[t]
  \centering
  \small
  \begin{tabular}{lccc}
   & Translation & Rotation & RMSE \\
   & (cm) & (deg) & (cm) \\
  \hline
  Our system           & $3.4 \pm 2.5$       & $1.4 \pm 0.7$ & $4.2$ \\
  RGB-D SLAM \cite{endres2012evaluation} & $9.6 \pm 5.7$ & $3.9 \pm 0.6$ & $11.2$ \\ 
  PTAM \cite{Klein07} & $5.0 \pm 2.4$ & $2.1 \pm 0.9$ & $ 5.6$ \\   
  \end{tabular}
  \caption{Absolute Trajectory Error. Translation and rotation mean error (mean~$\pm$~std) and translation RMSE of our system in comparison with \mbox{RGB-D} SLAM and PTAM.}
  \label{tab:GTerror}
  \end{table}
}

 \newcommand{\tabRPE}
{
  \begin{table}[t]
  \centering
  \small
  \begin{tabular}{lccc}
   & Translation & Rotation & RMSE \\
   & (cm) & (deg) & (cm) \\
  \hline
  Our system           & $5.1 \pm 3.3$       & $1.5 \pm 0.9$ & $6.1$  \\
  RGB-D SLAM \cite{endres2012evaluation} & $17.7 \pm 10.5$ & $4.9 \pm 2.3$ & $20.6$  \\ 
  PTAM \cite{Klein07} & $6.0 \pm 3.1$ & $1.1 \pm 0.5$ & $ 6.8$  \\   
  \end{tabular}
  \caption{Relative Pose Error . Translation and rotation mean error (mean~$\pm$~std) and translation RMSE of our system in comparison with \mbox{RGB-D} SLAM and PTAM.}
  \label{tab:RPE}
  \end{table}
}

\newcommand{\figGTtrajectory}
{
  \begin{figure}[b]
  \centering
  \includegraphics[width=\columnwidth]{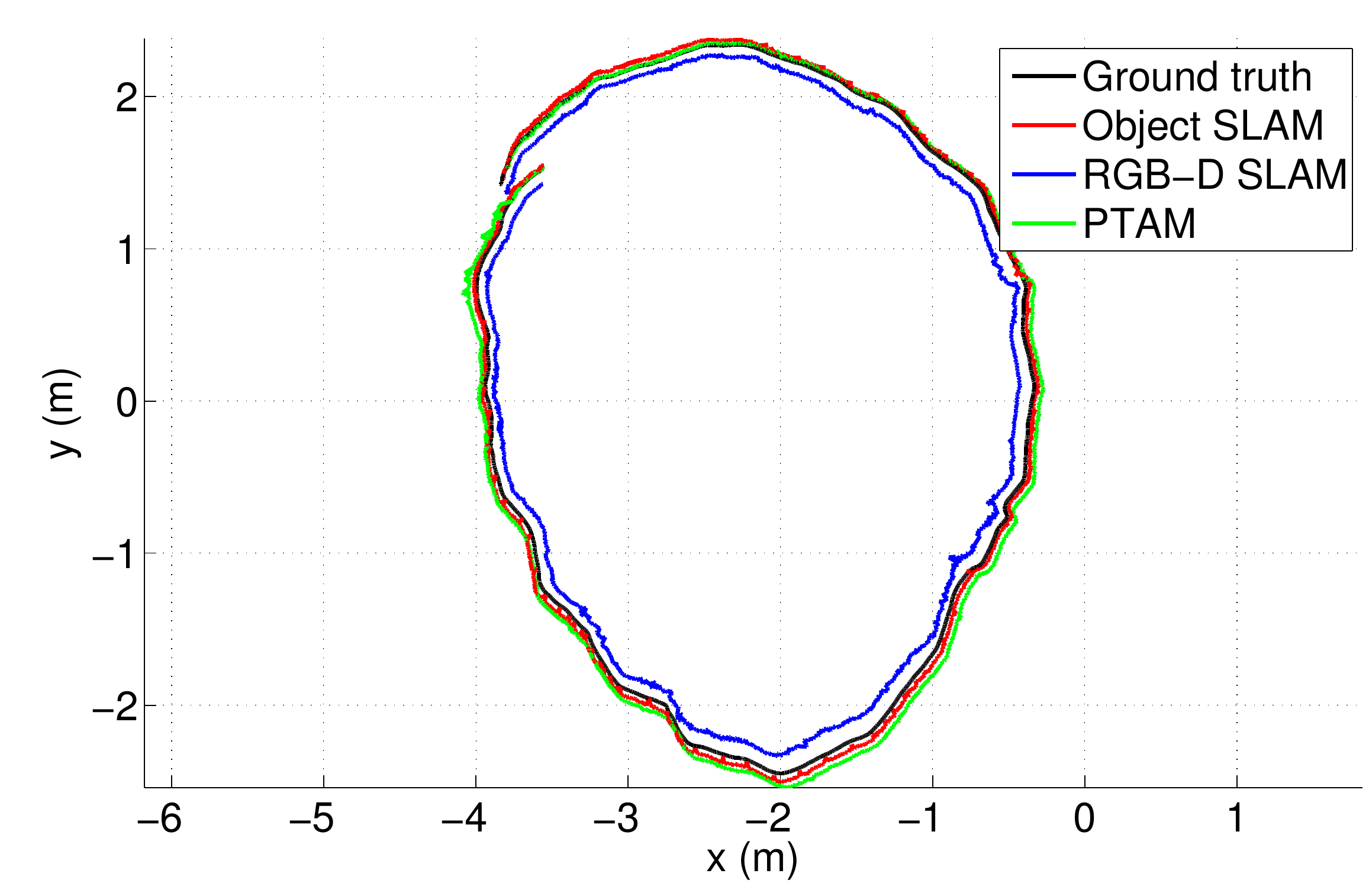}  
  \caption{Trajectory and scale estimated by our Object SLAM system
  in comparison with RGB-D SLAM \cite{endres2012evaluation} and PTAM \cite{Klein07} in the RGB-D SLAM Dataset. }
  \label{fig:GTtrajectory}
  \end{figure}
}

\newcommand{\figGTmap}
{
  \begin{figure}[!b]
  \centering
  \includegraphics[width=\columnwidth]{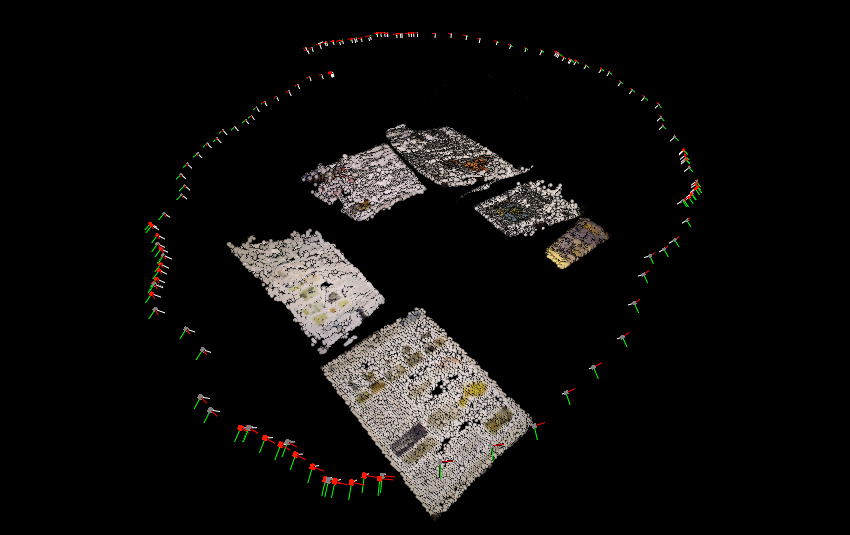}
  \caption{Map of objects created in the RGB-D SLAM sequence}
  \label{fig:GTmap} 
  \end{figure}
}

\newcommand{\figObjModel}
{
  \begin{figure}
  \centering
  \includegraphics[width=.9\columnwidth]{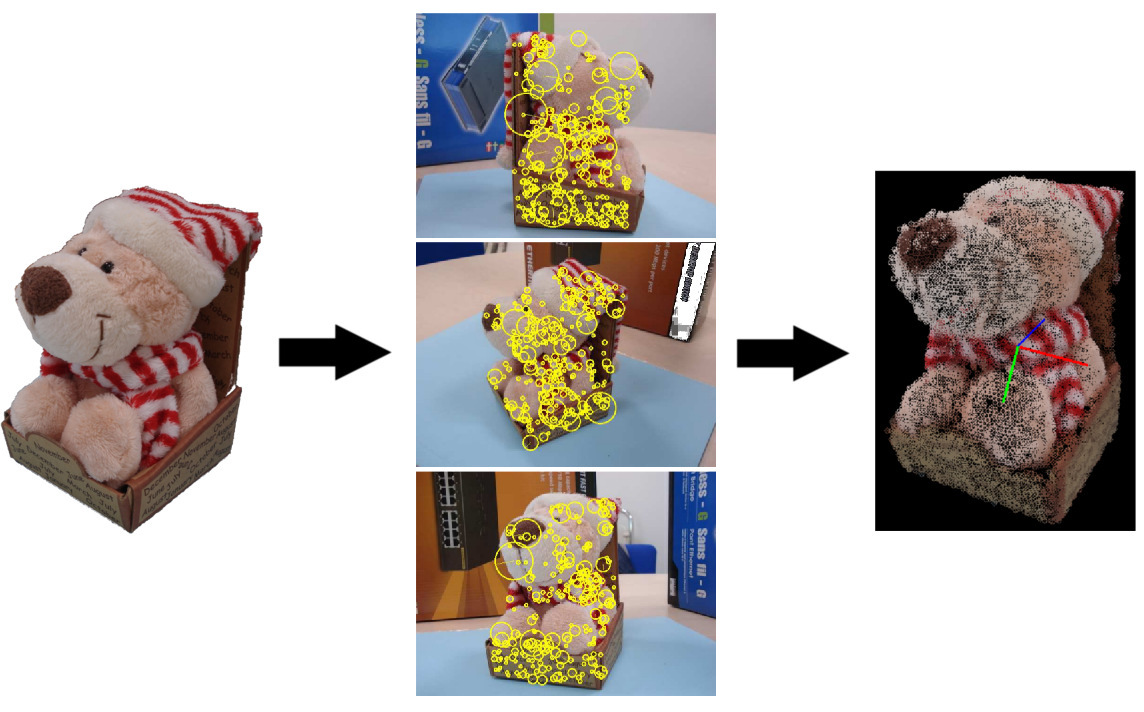}
  \caption{Objects are modeled with a point cloud obtained from multiple view
  geometry. }
  \label{fig:ObjModel}
  \end{figure}
}

\newcommand{\algObjectPriors}
{
  \begin{algorithm}[b]
    \vspace{2mm}
    \SetAlTitleFnt{} 
    \SetAlCapFnt{\small\centering} 
    \small
    \DontPrintSemicolon 
    \SetAlgoLined 

    \KwIn{Query image taken at position $\T_{WC_i}$}
    \KwIn{Set $\OO$ of objects previously observed}
    \KwOut{Set $\BB = \{ B^i_{1O}, B^i_{2O}, \dots \}$ of observed objects}
    
    $\BB \leftarrow \emptyset$\;
    
    \ForEach{$O_k \in \OO$}
    {
      Compute expected pose $\T^*_{C_iO_k}$\;
      Project $\PP_O$ on image with $\T^*_{C_iO_k}$\;
      Find new 2D-to-3D correspondences\;
      Estimate 3D pose to obtain observation $B^i_{O_k}$\;
      \If{\emph{pose found}}
      {
        Remove image features $\UU_i \in B^i_{O_k}$  from image\;
        $\BB \leftarrow \BB \cup \{ B^i_{O_k} \}$\;
      }
    }
    \Return{$\BB$}
    \vspace{2mm}
    
    \caption{Recognition of objects previously observed}
    \label{alg:ObjectPriors}
  \end{algorithm}
}

\newcommand{\algObjectGeneral}
{
  \begin{algorithm}[b]
    \vspace{2mm}
    \SetAlTitleFnt{} 
    \SetAlCapFnt{\small\centering} 
    \small
    \DontPrintSemicolon 
    \SetAlgoLined 

    \KwIn{Query image taken at position $\T_{WC_i}$}
    \KwOut{Set $\BB = \{ B^i_{1O}, B^i_{2O}, \dots \}$ of observed objects}
    
    $\BB \leftarrow \emptyset$\;
    
    Divide image feature into regions of interest\;
    \ForEach{\emph{region of interest}}
    {
      Query object database\;
      Compute correspondences with the top-10 candidates\;
    }
    \ForEach{\emph{object candidate}}
    {
      Join correspondences from all the regions\;
      Verify detection and obtain observation $B^i_O$\;
      \lIf{\emph{detection verified}}
      {
        $\BB \leftarrow \BB \cup \{ B^i_O \}$\;
      }
    }
    \Return{$\BB$}
    \vspace{2mm}
    
    \caption{General object recognition}
    \label{alg:ObjectGeneral}
  \end{algorithm}
}

\journal{Robotics and Autonomous Systems}

\begin{document}


\begin{frontmatter} 

\title{\LARGE \bf Real-time Monocular Object SLAM}

\author[unizar]{Dorian G\'alvez-L\'opez\corref{contribution}} \ead{dorian3d@gmail.com}
\author[unizar]{Marta Salas\corref{contribution}} \ead{msalasg@unizar.es}
\author[unizar]{Juan D. Tard\'os}  \ead{tardos@unizar.es}
\author[unizar]{J. M. M. Montiel}  \ead{josemari@unizar.es}

\cortext[contribution]
{
  These authors have contributed to the present work in equal parts.
}

\address[unizar]
{
  Instituto de Investigaci\'on en Ingenier\'ia de Arag\'on (I3A),
  Universidad de Zaragoza, 
  Mar\'ia de Luna 3, Zaragoza 50018, Spain.
}






\begin{abstract}

We present a real-time object-based SLAM system that leverages the largest object database to date. Our approach comprises two main components: 
1) a monocular SLAM algorithm that exploits object rigidity constraints to improve the map and find its real scale, and
2) a novel object recognition algorithm based on bags of binary words, which provides live detections with a database of 500 3D objects.
The two components work together and benefit each other: the SLAM algorithm accumulates information from the observations of the objects, anchors object features to especial map landmarks and sets constrains on the optimization. At the same time, objects partially or fully located within the map are used as a prior to guide the recognition algorithm, achieving higher recall. We evaluate our proposal on five real environments showing improvements on the accuracy of the map and efficiency with respect to other state-of-the-art techniques.
\end{abstract}


\end{frontmatter} 
\section{Introduction}
\label{sec:Introduction}

A robot that moves and operates in an environment needs to acquire live information about it in real time. This information can be obtained from Visual SLAM ({\em simultaneous localization and mapping}), a key component of many systems that allows mobile robots to create maps of their surroundings as they explore them, and to keep track of the location of themselves. Computed maps provide rich geometrical information useful for reliable camera location, but it is poor for describing the observed scene.  Recently, these maps have been augmented with objects to allow the robots to interact with the scene \cite{Salas13, CiveraIROS11, Castle2011524}.

To include objects in SLAM maps, these must be recognized in the images acquired
by the robot by computing a rigid-body 3D transformation. A vast research line has provided solutions to this problem \cite{Lutz13, Collet2011, Grundmann11, irosws11germandeli}, but it has been aside from visual SLAM.
 
Our aim in this paper is to approach object recognition and monocular object SLAM together, with a novel solution based on accumulating information over time
to compute more robust poses of objects and to keep them constantly located
in the scene.
To achieve this, we propose a novel object recognition algorithm
that provides detections of objects
as a keyframe-based visual SLAM algorithm builds a map.


Once an object is observed several times from different camera positions, those object features with several observations
are triangulated within the map as anchor points. 
Anchor points provide the location of the object within the map and set additional geometrical constraints in the bundle adjustment (BA) optimization. Since object models are at real scale, anchor points provide observations of the map scale.



Standard BA optimizes camera poses and map point locations and it is well known that it can only recover maps up to scale. In contrast, our algorithm optimizes the camera poses, the points as well as the anchor points, the objects and the scale, and as a result we have maps at real scale composed by objects. 

Our system relies on an object recognition algorithm that works on a single-image basis but takes advantage of the video sequence. It exploits the information collected by SLAM to treat previous observations as cues for the location of the objects in the current image. This allows to obtain faster and more repeatable detections that, in turn, provide more geometrical constraints to SLAM.  


\IGNORE
{
  To build them,
  before any insertion into the database, a large amount of samples of features 
  are required as training data. For that, the same SIFT features \cite{Lowe04}
  extracted from the object images that are going to be stored in the database 
  are commonly used \cite{Collet2011, irosws11germandeli}. 
  This forces to re-create the trees
  when a new set of objects are added into the database, which can be costly.
}
The novel object recognition algorithm we propose, based on bags of binary words
\cite{GalvezTRO12}, uses a static visual vocabulary that is independent of  
the number of objects, 
and models the entire appearance of the objects with 
ORB (\emph{oriented FAST and rotated BRIEF}) features
\cite{ORBiccv11}.
Poses of objects are found from 2D-to-3D correspondences that are refined by 
guided matching during a RANSAC-like step \cite{fischler1981random}. 
Our system performs a fast and reliable recognition of 3D objects with databases comprising up to 500 objects, while keeping the real-time constraints of a SLAM system.

Our work makes the following contributions:
\begin{enumerate}
\item We present a complete visual SLAM system that is able to insert real objects in the map and to refine their 3D pose over time by re-observation, with a single monocular camera.
\item We show the feasibility of storing hundreds of comprehensive 3D models in a single object database, composed of bags of binary words with direct and inverted indices. We also propose a novel technique to sample
putative correspondences in the verification stage.  
\item We propose a new SLAM back-end that includes the geometrical information provided by the objects 
into the map optimization to improve the accuracy of the map, the objects and their relative scale at each step.
\item We present results in real and independent datasets and comparisons 
with other systems. 
Our results proof that, by including objects, our monocular system can retrieve the real scale of the scene, 
and obtains more accurate results that PTAM~\cite{Klein07} and RGB-D SLAM~\cite{endres2012evaluation}, while keeping realtime performance 
(tracking takes $7.6$ ms, and recognition, around $200$ ms per image). Our results also demonstrate that the 
system is extremely robust against occasional wrong detections, avoiding map corruption.
\end{enumerate}

The paper is distributed as follows: 
Section~\ref{sec:RelatedWork} presents the related work of object SLAM
and object recognition.
Section~\ref{sec:Overview} gives an overview of our complete system.
Section~\ref{sec:SLAM} details the visual SLAM approach and the object insertion, and
Section~\ref{sec:ObjectRecognition}, the object recognition algorithm.
Section~\ref{sec:Evaluation} shows the experimental evaluation of our system,
and Section~\ref{sec:Conclusions} concludes the paper.



\section{Related work}
\label{sec:RelatedWork}
\figOverview

Object-augmented mapping has been previously approached by SLAM methods based on the extended Kalman filter \cite{CiveraIROS11, Castle2010ivc}. However, nowadays state-of-the-art monocular SLAM methods are based on keyframes, which create maps just with some selected video frames. As Strasdat\etal\cite{StrasdatICRA2010} proved, these systems are able to produce better results than filter based methods because they handle a great deal of points and produce larger accurate sparse point maps in real time at frame rate.

The work by Castle\etal\cite{Castle2011524} was one of the first ones that merged object recognition and monocular keyframe-based SLAM. After detecting an object in two frames, they compute its pose in the map. These objects are shown as augmented reality but, unlike our approach, they do not add the objects to the optimization. They built a database of 37 planar pictures described by SIFT features. Contrary to this system, restricted to planar objects, we can deal with objects with arbitrary 3D shapes.

Bao\etal\cite{Bao11,Bao12} were the first to present Semantic Structure from Montion (SSfM), which is a framework to jointly optimize cameras, points and objects. SLAM methods deal with the fact that the information proceeds from a video stream, thus the graph of points and keyframes is incremental, while SSfM processes all the frames at once. Moreover, recognition and reconstruction steps are separated and independent in \cite{Bao11}. However, on our algorithm, recognition and reconstruction take place at the same time since SLAM and object detection are fully integrated. The recognition method in \cite{Bao11} retrieves a bounding box of the object while our object detector retrieves a 6DoF pose. 

Along the same line, Fioraio\etal\cite{Fioraio13} presented a SLAM system that adds 3D objects in the maps when they are recognized with enough confidence, optimizing their pose together with the map by bundle adjustment. They build a database of 7 objects that are described by 3D features that are acquired at several scales with a RGB-D camera, creating an independent index for each scale. The recognition is performed by finding 3D-to-3D putative correspondences that are filtered by a RANSAC-based algorithm. Although they are able to build room-sized maps with a few objects, their system does not run in real time. In comparison, our system improves scalability and execution time by using binary 2D features and a single index structure that can deal with all the keypoint scales at the same time.

Salas-Moreno\etal\cite{Salas13} presented one of the most recent visual SLAM systems that combines RGB-D map building and object recognition. They represent the map with a graph in which nodes store position of cameras or objects, and enhance the pose of all of them when the overall graph is optimized. A database of objects is built beforehand with KinectFusion~\cite{newcombe2011kinectfusion}, describing their geometry with point pair features \cite{Drost10}. These are indexed by a hash table, and the recognition is performed by computing a large number of candidate rigid-body transformations that emit votes in a Hough space. Hough voting is a popular technique for object detection with RGB-D data \cite{Wang13, Barinova12,Sun2010}, but its scalability to hundreds of objects is not clear. In fact, Salas-Moreno\etal\cite{Salas13} achieve real-time execution by exploiting GPU computation, but they show results with just 4 objects. In our work, we show results at high frequency with up to 500 objects, computed on a CPU from a monocular camera.


Regarding object recognition, our proposal follows the line of research consisting in finding matches of local features between an image and an object model.
Sivic \& Zisserman \cite{Sivic03}
presented a visual vocabulary to match 2D images in large collections. 
They proposed to cluster the descriptor space of image features with {\em k}-means
to quantize features and represent images with numerical vectors, 
denoted as bags of words, enabling quick comparisons. 
On the other hand, Lowe~\cite{Lowe04} popularized an approach
based on directly matching SIFT features between query and model 2D images.
Matching features requires to compute the descriptor distance 
between large sets of features, which can be very time-consuming. 
To speed up this process, he proposed the \emph{best-bin-first} technique to 
find approximate neighbors with a {\em k}-d tree. 
Both visual vocabularies and {\em k}-d trees were later generalized for
matching large sets of images in real time. 
Nister \& Stewenius \cite{Nister06} 
presented a hierarchical visual vocabulary tree built on 
MSER (\emph{maximally stable extremal regions}) \cite{matas2004robust} and SIFT features with which
yielded fast detections with a dataset of 40000 images. 
Muja \& Lowe \cite{muja_flann_2009} presented a method to automatically configure 
a set of {\em k}-d trees to best fit the feature data to match.

To fully recover the pose of an object from a single image, it is necessary
to incorporate 3D information to the models. 
Gordon \& Lowe \cite{gordon2006} started to create 3D point cloud models, 
recovering the object structure by applying 
structure from motion techniques.
The pose could be then retrieved by solving the perspective-{\em n}-problem
\cite{fischler1981random} from 2D-to-3D correspondences.
This has been the basis of a lot of recent object recognition approaches
\cite{Collet2011, Sattler2011, irosws11germandeli, Grundmann11, hsiao2010making}.
For example, Collet\etal\cite{Collet2011} build 3D models for 79 objects and 
use the training images of the objects to build a set of 
{\em k}-d trees to index their SIFT features and do direct matching. 
To enhance the detection of small objects and avoid the background, 
they run the recognition on small sets of SIFT features that are close in the 
query image, merging the detections later if the object poses overlap. 
Although we also divide the query features into regions, we merge the 
2D-to-3D correspondences before computing any pose. 
This prevents from missing detections in the cases of
oversegmented regions with few correspondences.
In a similar way, Pangercic\etal\cite{irosws11germandeli} 
create a database of 50 3D objects represented with a SIFT vocabulary tree, 
trained with the same object images. They rely on a RGB-D camera to segment out
the background.

The diverse discretization levels of trees allow to compute feature
correspondences in several manners. 
For example, Hsiao\etal\cite{hsiao2010making}
discretize the SIFT descriptor space in a hierarchical manner
to create a 3-level tree. They show the benefits of computing
feature matches at all the levels and not only the finest one, 
obtaining more putative correspondences that increase the object recognition rate. 
However, an excess of correspondences may overburden the pose recovery
stage, leading to a large execution time.
In contrast, Sattler's\etal\cite{Sattler2011} approach retrieves correspondences
only from those features that lie in the same visual word, 
but this may miss correct pairs of points that do not share the visual word due
to discretization error. 
In our work, we use a direct index \cite{GalvezTRO12} to compute correspondences 
between features that lie in the same tree node at a coarse discretization level.
This provides a balanced trade-off between amount of corresponding points
and execution time.

All these works use SIFT or SURF features, which are described with vectors
of 64 or 128 float values, and train matching trees with the same images
with which the objects are modeled, which forces them to recreate the trees
when new objects are added to the database.
Rublee\etal\cite{ORBiccv11} 
presented ORB features, which are binary and compact (256 bit length descriptor), 
and provide a distinctiveness similar to that of SIFT and SURF \cite{Miksik12}. 
Furthermore, visual vocabularies of binary words created from independent data
and that do not need reconstruction are suitable to index
large collections of images \cite{GalvezTRO12}. 
We show in this work the viability of a single independent vocabulary of ORB features 
to recognize 3D objects with large databases (up to 500 objects) in real time
(around 200 ms/image).

\section{System overview}
\label{sec:Overview}


Our system builds a 3D map composed by camera poses, points {\em and objects}, as illustrated by Figure~\ref{fig:Overview}.
We make use of the front-end of the {\em Parallel Tracking and Mapping} (PTAM) algorithm \cite{Klein07} to track the camera motion, and add two new parallel processes to perform object recognition and object insertion in the map.
Our system also includes a completely redesigned back-end based on g2o \cite{Kummerle11} that performs a {\em joint} SLAM optimization of keyframe poses, map points, objects and map scale. 


The \emph{SLAM tracking} processes all the video frames to 
compute the pose of the camera at every time step
with an unknown map scale.
When a frame provides distinctive geometrical information, it is inserted in
the map as a \emph{keyframe} together with new map points.

Simultaneously, object recognition is performed on as many frames as possible to
search for known objects stored in an \emph{object model database}.
If there is available information of the location of objects, given by 
both the SLAM map and previous recognitions,
this is exploited to guide the detection in the current image.  
A successful detection provides an \emph{observation} of an object instance.


Regardless of the recognition algorithm used, a detection obtained from a single image may be spurious or inaccurate.
To avoid these problems,
instead of placing an object in the map after a first recognition, 
we insert it in the SLAM map after \emph{accumulating} consistent observations 
over time. 
The information given by all the observations is used to triangulate
the object points, and hence the pose of the object 
inside the SLAM map.
The resulting points are inserted in the 3D map as \emph{anchor points}, 
and the cameras that observed them, as \emph{semantic keyframes}. 
These keyframes are not selected because of a geometrical criteria,
but because they contain relevant semantic information. 
The frames of the observations that do not provide parallax or distinctive 
geometrical information are discarded. 
Each triangulation provides us with an estimate of the map scale 
which we use to globally optimize it.  




\section{Object-aware SLAM}
\label{sec:SLAM}

\subsection{Object insertion within the map}
\label{sec:insertion}

The recognitions of objects in single images yielded by our algorithm 
are used to insert those objects in the SLAM map. 
To robustly place them, instead of relying on a single detection,
we accumulate several of them until we have enough geometrical information
to compute a robust 3D pose.
This process is depicted by Figure \ref{fig:triangulation} and explained next.

\figTriangulation

The object detector described in Section \ref{sec:ObjectRecognition} searches for 
objects in as many frames as possible, whereas SLAM uses them to track the camera,
so that the pose 
$\T_{WC_i} = \left[ \R_{WC_i} \, | \, s \, \mathbf{t}_{WC_i} \right]$ 
of each camera $i$ is known, with a map scale $s$ that is initially unknown.
A successful recognition of an object model $O$ returns a transformation $\T_{C_iO}$
from the camera to the object frame. 
Since multiple physical instances of the same object model may exist, we check  
to which instance this detection belongs. We do so by computing a hypothesis
of the global pose $\Ts_{WO} = \T_{WC_i} \, \T_{C_iO}$ of the detected object in the world,
and checking for overlap with the rest of the objects of the same model that
had been previously observed or are already in the map.
Note that this operation is valid only if we already have an estimate of the map
scale $s$. Otherwise, we assume that consecutive detections of the same model come
from the same real object. After this, we determine the detection of 
object $O$ is an observation of $O_k$, the $k$-th instance of model $O$. 
If there is no overlap with any object observed before, we just create a new
instance.  

An observation 
$B^i_{O_k} = \quadruple{\T_{WC_i}}{\T_{C_i O_k}}{\XX_O}{\UU_i}$
yields a set of correspondences between some 3D points of 
model $O$, $\XX_O$, and 2D points of the image taken by camera $C_i$, $\UU_i$. 
For each 
correspondence $\pair{\xx_O}{\uu_i} \in \pair{\XX_O}{\UU_i}$,
if the parallax with respect to the rest of observations of $\xx_O$ of the same 
object instance is not significant enough, the corresponding pair is discarded.
If an object observation does not offer parallax or new points, 
it is completely disregarded. 

The observations of an object instance are accumulated until the following conditions hold: 
1) at least 5 different object points $\xx_O$ are observed from two different
positions, 
2) with at least 3 degrees of parallax between the cameras, and 
3) showing no alignment and a good geometrical conditioning. 
The points are triangulated in the world frame ($\xx_W$) and the pairs
$\pair{\xx_O}{\xx_W}$ are inserted in the map 
as anchor points. The frames that offered parallax are also inserted as semantic keyframes.

Anchor points play a decisive role on the object SLAM, because they provide the location of the object within the map and set additional geometrical constraints in the BA enabling the map scale estimation. For this reason, anchor points have a different treatment than map points: they are not discarded by the maintenance algorithm of PTAM, are updated using new object observations only and are propagated among the keyframes of the map by using matching cross correlation in a $3\times 3$ pixel region defined around the projected anchor point in the target keyframe. The patches for the correlation are extracted from the semantic keyframes and warped in order to compensate scale, rotation and foreshortening by means of a homography. 


By triangulating object points from several observations, we provide a more robust
3D pose than relying on a single detection. Furthermore, this operation is necessary
to find the map scale $s$ if our only source of data is a monocular camera.
Since we aim for $\T^{-1}_{WC_i} \, \xx_W = \, \T_{C_i O_k} \, \xx_O$ for each point, 
an estimate of the map scale is given by each triangulated object instance: 
\begin{equation}
\begin{split}
&s_{O_k} = \\
&\argmin_s \sum_{i} \sum_{\pair{\xx_O}{\xx_W}}
  \left\lVert
    \R^t_{WC_i} \xx_W - \R_{C_iO_k} \xx_O - \tr_{C_iO_k} - s \, \R^t_{WC_i} \tr_{WC_i}  
  \right\rVert^2.
\label{eq:sok}
\end{split}
\end{equation}

To insert the object in the map, we must compute its pose $\T_{WO_k}$ in the map world frame.
After the first triangulation of an object, we compute an initial pose which is used as a prior in the subsequent SLAM optimization.  
The object pose prior  is computed by composing the information provided by the SLAM and the object detector by means of equation \eqref{eq:TwoComposition}. This composition is shown in Figure~\ref{fig:ScaleEstimation}. 
\begin{equation}
T_{WO_k} = \left[  \R_{WC_i}\R_{C_iO_k} \, | \, \R_{WC_i} \tr_{C_iO_k} + \hat{s} \, \tr_{WC_i} \right].
\label{eq:TwoComposition}
\end{equation}
The pose of the semantic keyframe $T_{WC_i}$ and  the pose of the object with respect to the camera $T_{C_iO_k}$ corresponds to the information of the last observation $B^i_{O_k}$.
The scale $\hat{s}$ used is either the scale estimate $s_{O_k}$ computed above, or the map scale $s$ if we already had a previous estimate that had been refined in the optimization stage.  

The anchor points, the semantic keyframes and the object pose priors 
produced by each triangulation are then included 
in the optimization stage of the SLAM mapping algorithm to obtain
more accurate values during the SLAM execution.

\figScaleEstimation


\subsection{Object SLAM optimization}
\label{sec:optimization}

\figBayesianNetworks

In standard keyframe-based SLAM, a sparse map of points $\XX_{W}$ and the camera location of selected keyframes $T_{WC_i}$  are estimated by means of a joint bundle adjustment (BA).
Figure~\ref{fig:BayesianSlam} shows a Bayesian network representing the estimation problem structure.

The BA minimizes \emph{the map reprojection error}, $\ee_{ji}$, between the 
$j$-th map point observed by the $i$-th keyframe and the corresponding 
measurement $\uu_{ji}=\left(u_{ji},v_{ji}\right)^t$:
\begin{equation}
  \ee_{ji} = \left(\begin{array}{c} u_{ji}\\ v_{ji} \end{array}\right)
    - \text{CamProj} \left(T^{-1}_{WC_i} \, \xx_{jW}\right).
\label{eq:reprojection error}
\end{equation}

\newcommand{\vecB}[2]{\left(\begin{array}{c} #1 \\ #2 \end{array}\right)}
\newcommand{\vecC}[3]{\left(\begin{array}{c} #1 \\ #2 \\ #3 \end{array}\right)}
\newcommand{\proj}{\text{CamProj}}
\newcommand{\diagB}[2]{\left[\begin{matrix} #1 & 0 \\ 0 & #2 \end{matrix}\right]}
\newcommand*\rfrac[2]{{}^{#1}\!/_{#2}}

The point in the camera frame $\xx_{jC_{i}} =T^{-1}_{WC_i} \xx_{jW}$ is projected 
onto the image plane through the projection function 
$\textit{CamProj}: \mathbb{R}^3 \mapsto \mathbb{R}^2$ defined by 
Devernay \& Faugeras \cite{Devernay01}.

\IGNORE
{
  The point in the camera frame $\xx_{jC_{i}} =T^{-1}_{WC_i} \xx_{jW}$ is projected 
  onto the image plane through the projection function 
  $\textit{CamProj}: \mathbb{R}^3 \mapsto \mathbb{R}^2$ defined by \cite{Devernay01}:
  \begin{eqnarray}
    \proj \vecC{x_{ji}}{y_{ji}}{z_{ji}}  = 
      \vecB{u_0}{v_0} + \diagB{f_u}{f_v} \frac{r'}{r \, z_{ji}}
      \vecB{x_{ji}}{y_{ji}}, \\
    r = \sqrt{\frac{x_{ji}^2+y_{ji}^2}{z_{ji}^2}}, ~ 
    r'= \frac1{\omega}\arctan\left(2r~\tan\frac{\omega}{2} \right).
  \end{eqnarray}
  
  The intrinsic parameter of the camera are the focal length $(f_u,f_v)$, the principal point $(u_o,v_o)^t$ and the distortion $(w)$.
}

Our goal is to include in the estimation the constraints given by the 
triangulation of a set of $K$ objects, and then obtain optimized estimates of
the map cameras, points, objects and scale.
A triangulated object instance $R_{O_k} = \triple{T_{WO_k}}{\XX_O}{\XX_W}$ comprises
a set of 3D points $\XX_{O}$ in the object frame and its corresponding anchored
landmarks in the map, with coordinates $\XX_W$ in the SLAM world frame.
It is well known that from monocular sequences, the scene scale $s$ is unobservable. However, if some map points correspond to a known size scene object (anchor points) the resulting geometrical constraints allow to estimate the SLAM scale. Figure~\ref{fig:BayesianObjectSlam} shows the new estimation problem structure after including anchor points. New nodes are added to the Bayesian network because new parameters have to be estimated: object locations $\T_{WO_k}$ and map scale $s$, which is observable now, being a single value for all the inserted objects.

Each anchor point $\pair{\xx_O}{\xx_W} \in \pair{\XX_O}{\XX_W}$ sets a new constraint, the \emph{object alignment error}, $\mathbf{a}_{jk}$, defined as the difference between 
the positions of the point when both measures are translated into the same object frame of reference:
\begin{equation}
  \aaa_{jk} = \xx_{jO_k} - s \, \R_{WO_k}^t \xx_{jW} +  \R_{WO_k}^t \tr_{WO_k}.
  \label{eq:aligment error}
\end{equation}


We propose a BA to iteratively estimate the scale, the map points, the cameras and the objects, by minimizing a robust objective function combining the reprojection error \eqref{eq:reprojection error} and the object alignment error \eqref{eq:aligment error}:
\begin{equation}
  \hat{\mathbf{p}} = \argmin_{\mathbf{p}} \sum_{i=1}^N\sum_{j \in \SSS_i} H\left(\ee_{ji}^t \Omega^C \ee_{ji} \right) +
  \sum_{k=1}^{K}\sum_{j \in \SSS_k} H\left(\aaa_{jk}^t \Omega^O \aaa_{jk} \right),
\label{eq:BA}
\end{equation}
%
where $N$ is the total number of keyframes, $\SSS_{i}$ is the subset of map points seen by the $i$-th camera, $\SSS_k$ is the subset of anchor points of the $k$-th object instance, $\Omega^C$ and $\Omega^O$ are the information matrices of the reprojection error and the alignment error respectively. Errors are supposed uncorrelated and follow a Gaussian distribution, thus the covariance matrices are diagonal. Regarding $\Omega^C$, it is a 2x2 matrix and the measurement error is $\sigma^2 = 2^{2l}$, where $l$ is the level of the pyramid in which the feature was extracted. Similarly, $\Omega^O$ is a 3x3 matrix with measurement error $\sigma^2 = 0.01^2$. $H(\cdot)$ is the Huber robust influence function \cite{Hartley2004}:
\begin{equation}
H(x) = \left\{ 
  \begin{array}{ll} 
    x & \text{if } \lvert x \rvert < \delta^2, \\ 
     2  \delta x^{ \frac{1}{2}} - \delta^2 & \text{otherwise.}
  \end{array} 
  \right.
\label{eq:huber}
\end{equation}
Here, $\delta^2$ corresponds to the $\chi^2_{0.05}(r)$ distribution, being $r$ the number of degrees of freedom. The $\delta^2$ value for the reprojection error \eqref{eq:reprojection error} is $\chi^2_{0.05}(2) = 5.991$ and for the alignment error \eqref{eq:aligment error} is $\chi^2_{0.05}(3) = 7.815$.

The optimization vector is 
\begin{equation}
 \mathbf{p}=\left( s, ~ \boldsymbol\nu_{WC_2}, \dots, \boldsymbol\nu_{WC_N}, ~ \boldsymbol\nu_{WO_1}, \dots, \boldsymbol\nu_{WO_K}, ~ \xx_{1W}, \dots, \xx_{MW} \right),
\end{equation}
where $s$ is the map scale 
and $\boldsymbol\nu$ represents a transformation parametrized as a 6 component vector (rotation and translation) of the SE(3) Lie group.


While the camera is exploring the scene, new keyframes are inserted in the map. To compute a prior pose for the new keyframes a sliding window is applied to the keyframes of the map and only four neighbor keyframes of the new keyframe and all the visible points are included in a BA, minimizing the reprojection error eq~\eqref{eq:reprojection error}. The object pose priors are computed as previously explained in this section, following eq~\eqref{eq:TwoComposition}. Global BA \eqref{eq:BA}, including all the cameras, points and objects, is performed every time a new keyframe or a new object is inserted within the map.

\section{3D Object recognition with large databases}
\label{sec:ObjectRecognition}

The object recognition requires a visual vocabulary built from an independent set of images, and a database of models that is created offline.
Then, the recognition process is executed online in real time,
performing two main steps on a query image
taken at position $\T_{WC_i}$: 
\emph{detection} of several model candidates that fit the image features, 
and \emph{verification} of the candidates by computing a rigid body transformation
between the camera and the objects.
The candidates are obtained either by querying all the models in a database 
based on bag of words, or by taking advantage of previous known locations
of objects.
The verification step makes use of 2D-to-3D correspondences between image and
object model points to find the object pose in the image $\T_{C_iO}$.
The result are observations $B^i_O = \quadruple{\T_{WC_i}}{\T_{C_i O}}{\XX_O}{\UU_i}$
of the object models $O$ recognized. 
The SLAM algorithm associates then these results to their corresponding 
object instances $O_k$ taking the pose of the current camera into account
(Section~\ref{sec:insertion}).


\subsection{Object models}
\label{sec:ObjectModels}

Our object models are composed of a set 3D points associated to ORB 
descriptors and an appearance bag-of-words representation for the complete
object. 
ORB features are computational efficient because they describe image 
patches with strings of $256$ bits. 

Each object model $O$ is created offline from a set of 
training images taken
from different points of view of the object. 
We use the Bundler and PMVS2 software 
\cite{snavely2006photo, Furu:2009:MVS} 
to run bundle adjustment
on these images 
and to obtain a dense 3D point cloud of the object $\PP_O$, 
as shown by Figure~\ref{fig:ObjModel}.
We keep only those points that consistently appear in at least 3 images.
Since objects can appear at any scale and point of view during recognition, 
we associate 
each 3D point to several ORB descriptors 
extracted at different scale levels (up to 2 octaves) and from several training images.

If the point of view of the training images hardly differs, we may obtain
3D points with very similar descriptors that add little distinctiveness.
To avoid over-representation, we convert features into visual words
and keep the average descriptor per 3D point and visual word \cite{Sattler2011}.
Finally, an appearance-based representation of the object is obtained by 
converting the surviving binary features of all its views into a bag-of-words 
vector with a visual vocabulary. 
This model provides information of all the object surface, so that 
a single comparison yields a similarity measurement independently of
the viewpoint and the scale of the object in the query image. 

\figObjModel

\subsection{Object model database}

The object models are indexed in a 
database  composed of a visual vocabulary, an inverted index
and a direct index \cite{GalvezTRO12}. 
The visual vocabulary consists in a tree with binary nodes
that is created by hierarchical clustering of training ORB descriptors.
The leaves of the tree compose the words of the visual vocabulary. 
We used  12M descriptors obtained from 
30607 independent images from Caltech-256 \cite{griffinHolubPerona}
to build a vocabulary with $k=32$ branches and $L=3$ depth levels,
which yields 33K words.
When an ORB feature is given, its descriptor vector traverses
the tree from the root to the leaves, selecting at each level the node
which minimizes the Hamming distance, and obtaining the final
leaf as word. 
By concatenating the equivalent words of a set of ORB features,
we obtain 
a bag-of-words vector, whose entries
are weighted
with the term frequency -- inverse document frequency ({\em tf-idf})
value, and normalized with the $L_1$-norm.
This weight is higher for words with fewer occurrences in the training
images, since they are expected to be more discriminative.  

The inverted index stores for each word in the vocabulary the objects 
where it is present, along with its weight in that object. 
When a query image is given, 
this structure provides
fast access to
the common words between the query bag-of-words vector and the model one.
The direct index stores for each object model the 
tree nodes
it  contains and the associated ORB features.
This is used to discriminate those features that are likely to match
when 2D-to-3D correspondences are required in the verification stage.
We can increase the amount of correspondences if we use
the direct index to 
store nodes at other tree levels (coarser  discretization levels), 
with little impact on the execution time \cite{GalvezTRO12}.
In this work, we store nodes at the first discretization level 
of the vocabulary tree.

\subsection{Prior knowledge to obtain object candidates}
\label{sec:ObjectRecognition:Prior}

\algObjectPriors

The first method to obtain detection candidates arises from those objects that
have been previously observed or inserted in the map.
Detecting objects that are already in the map is useful because we can find
new points that were not anchored yet to landmarks. Inserting them help
optimize the pose of the object.
The process is described in Algorithm~\ref{alg:ObjectPriors}.

We have two sources of information of observed objects: 
triangulated objects inserted in the SLAM map, with 
optimized poses, and non-triangulated accumulated observations. 
From these, we can estimate the expected pose $\Ts_{C_iO_k}$ of each 
object instance in the current image if the map scale $s$ 
has been estimated. 
If it has not, we assume that $\Ts_{C_iO_k}$ is the same than the
last computed transformation $\T_{C_jO_k}$ if it was obtained recently 
(up to 2 seconds ago). The transformation $\Ts_{C_iO_k}$ is computed as:
\begin{equation}
\Ts_{C_iO_k} = 
  \left\{ \begin{array}{ll}
    \T^{-1}_{WC_i} \, \T_{WO_k}  & \text{if $O_k$ in the map,} \\
    \T^{-1}_{WC_i} \, \T_{WC_j} \, \T_{C_jO_k}  &  
      \text{if $O_k$ not in the map but $s$ known,} \\
    \T_{C_jO_k} &  \text{if $s$ unkown and $i - j \leq 2$ secs.}
  \end{array} \right.
\end{equation}

To obtain object candidates, we first extract ORB features from the query image.
For every object instance $O_k$ of which we can compute an expected pose 
$\Ts_{C_iO_k}$,
if it is visible from the current camera $C_i$, 
we project the object model 3D points $\PP_O$ on the image to look for 
correspondences following the same procedure 
explained in Section~\ref{sec:ObjectVerification}.
We estimate a 3D pose from these correspondences by solving the 
perspective-{\em n}-problem \cite{fischler1981random}. 
If it is successful, the utilized 2D features are removed from the image
and an object observation $B^i_{O_k}$ is produced.


\subsection{General retrieval of object candidates}

\algObjectGeneral

After trying to recognize objects previously seen,
the general retrieval of object candidates is performed
to find new detections. This is described in Algorithm~\ref{alg:ObjectGeneral}.

\figClusteringQuery

Objects can appear at any distance from the camera, so the detection of candidates
should be robust against scale changes. Sliding window techniques \cite{deSande11, Lampert08} 
are a common approach to face this difficulty by
searching variant size areas of the image repeatedly.  
In contrast, we rely on dividing the image into regions of interest  to perform
detections in small areas, merging results if necessary.
We run the Quick Shift algorithm \cite{vedaldi08quick}
on the ORB features of the query image to group together those
that are close in the 2D coordinate space
to obtain regions of interest.
Quick Shift is a fast non-parametric clustering algorithm that separates  
N-dimensional data into an automatically chosen number of clusters. In our case, 
each resulting 2D cluster defines a region of interest. 

The ORB features of each region of interest are converted into a bag-of-words
vector $\vv$ that queries the object database individually. The dissimilarity 
between the query vector and each of the object models $\ww$ in the database
is measured with a score based on the Kullback-Leibler (KL) divergence \cite{kullback1951information}. 
This score $s_{\kl}$ benefits from the inverted index since its computation
requires operations between words in common only, while the properties of the 
KL divergence are kept (\textit{cf}. \ref{apx:kl}):  
\begin{equation}
  s_{\kl}(\vv, \ww) = \sum_{v_i \neq 0 \, \wedge \, w_i \neq 0} v_i \log \frac{\varepsilon}{w_i},
\end{equation}
where $\varepsilon$ is a positive value close to $0$. 
In Section~\ref{sec:EvaluationDesktop} we compare the performance of the 
KL divergence with other popular metrics.
The sparsity of vectors $\vv$ and $\ww$ highly differs, since an object model
may comprise thousands of words whereas an image region just a dozen. 
This seems a difficulty for the retrieval of the correct model, however the
bag-of-words scheme can handle this situation because it can compare vectors
independently of their number of words. If a single word in a region
matches a model, its tf-idf weight can already produce an object candidate
with a  $s_{\kl}$ score. Perceptual aliasing may yield wrong candidates, 
but the correct model is expected to produce more correct word matches, lowering
its dissimilarity value. As a result, the correct object model is likely to be
retrieved even if only a few words in common are found.
Thus, the top-10 object models that offer the lowest dissimilarity score with
vector $\vv$ are selected as detection candidates for each region of interest.

Then, correspondences between the 2D image points and the 3D model points are
obtained. This operation is sped up by using the direct index to filter out
unlikely correspondences \cite{GalvezTRO12}. 
Our segmentation into regions bears a resemblance with other approaches. 
For example, MOPED \cite{Collet2011} tries to recognize an object only with
the correspondences obtained in each region, merging later the detections
if their poses overlap. In contrast, we merge 
the correspondences of
each region of interest according to their associated object model
before computing any pose.  
This prevents from missing detections due to oversegmented regions with few
correspondences.  
We finally find the pose of the object candidates, or discard them, in the 
object verification stage.

\figClusteringVan

Figure \ref{fig:ClusteringQuery} shows an example of how regions of interest
can help find small objects: a carton bottle is searched for from a database
with 500 object models. In Figure \ref{fig:ClusteringQuery-noclusters}, the
query image is not divided into regions and the entire image queries the 
database. As a result, the background makes the correct model appear as the 7th
best candidate, and prevents from obtaining correct correspondences, missing
the detection. On the other hand, as shown by Figure \ref{fig:ClusteringQuery-clusters},
when small regions are considered, we find the correct model as the 1st
candidate of its region, obtaining a better inlier ratio of correspondences
(6 out of 9), being able to verify the recognition. 
Thus, in addition to detect small objects, regions of interest are helpful
to establish better point correspondences.
Figure \ref{fig:ClusteringVan} shows an example in which a toy van is
divided into two regions. Since region correspondences are finally merged, 
it can be correctly recognized.


\subsection{Object verification and pose estimation }
\label{sec:ObjectVerification}

After obtaining putative 2D-to-3D correspondences between the query image 
and the object candidates, we try to verify and find the pose of each object
by iteratively selecting random subsets of correspondences and solving the 
perspective-\textit{n}-point problem (PnP) \cite{fischler1981random}. 
Plenty of algorithms based on random sample consensus (RANSAC) \cite{fischler1981random} 
exist to achieve this. For example, 
progressive sample consensus (PROSAC) \cite{chum2005matching} 
arranges the correspondences according to their distance in the 
descriptor space. Then, ordered permutations of low distance are selected 
as subsets for a
parametrized number of tries, after which the algorithm falls back to
RANSAC. This is usually much faster than RANSAC when the pose can be found.
However, in the presence of mismatching correspondences with low descriptor
distance (e.g. due to perceptual aliasing), PROSAC may spend several tries 
trying subsets with outliers. 
Since we set a low number of maximum iterations (50) to limit the impact in the
execution time, we propose a variation of PROSAC that eases the rigidity
of the fixed permutations and is thus more flexible when there are low-distance
mismatches. We coined it \emph{distance sample consensus} (DISAC) and consists in
drawing correspondences $c_j$ from the set of correspondences 
$\mathcal{C} = \left\{ c_1, \dots, c_n \right\}$ 
with a probability inversely proportional to its Hamming distance $h$:
\begin{equation}
P(c_j) =  \frac1{h(c_j) \displaystyle \sum_{k=1}^n \frac1{h(c_k)}}.
\end{equation}
To avoid numerical inconsistencies, we set $h(c_i) = 1$ when the distance
is exactly $0$.
Now, in the case of outliers, there is a non-zero probability to avoid them
even in the first iterations of DISAC. 

\figRansacSteps

If a 3D pose is found with the selected subset of correspondences, 
we try to refine it by selecting additional correspondences that
were not given by the direct index.
For that, we project the object model 3D points $\PP_O$ on the query image,
obtaining a set of 2D points.
For each visible point $\xx_O \in \PP_O$, we search a 
$7\times7$ patch centered at its projection for a matching ORB image feature $\uu$.
We consider them to match if any of the ORB descriptors associated to the 
corresponding 3D point is at a Hamming distance lower than $50$ units, 
which assures a low ratio of mismatches \cite{ORBiccv11}.
If new correspondences are found, we compute a new refined pose $\hat{T}_{C_iO}$, 
as shown by Figure~\ref{fig:RansacSteps}. 
We measure the quality of a pose $\hat{T}_{C_iO}$ with a reformulation
of Torr \& Zisserman's M-estimator \cite{Torr2000} on the reprojection error
that also takes into account the number of inlier correspondences:
\begin{equation}
  s_\text{DISAC} = \sum_{\pair{\uu}{\xx_O}} \max 
    \left( 
      0, ~ \mu_e -
      \left\lVert 
        \uu - \text{CamProj}(\hat{T}_{C_iO} \, \xx_O) 
      \right\rVert
    \right),
\end{equation} 
where \textit{CamProj} is the projection function presented in Section~\ref{sec:optimization}, and $\mu_e$, a threshold in the reprojection error set to $3$px.

We keep the refined transformation of maximum $s_\text{DISAC}$ score of all
the DISAC samples for each object candidate, if any, 
verifying the recognition and finding the transformation $\T_{C_iO}$ 
between the current camera and the object.
This, together with the collection of 2D-to-3D correspondences,
composes the object observation 
$B^i_{O_k} = \quadruple{\T_{WC_i}}{\T_{C_i O_k}}{\XX_O}{\UU_i}$
that feeds the SLAM algorithm.

%
%
%

\section{Experimental evaluation}
\label{sec:Evaluation}

Our system has been implemented in C++, as modules of the Robot Operating System (ROS) 
\cite{quigley2009ros}, exploiting parallelization with OpenMP \cite{Dagum98OpenMP}
in the object candidate detection and verification steps.
All the tests were done on a Intel Core i7 @ 2.67GHz PC. 

We evaluate our system in five different datasets with sets of from 7 to 500 objects: 
the \emph{Desktop} dataset, used for testing purposes;
one of the sequences of the 
\emph{RGB-D SLAM Dataset} \cite{sturm12iros}, which
provides ground truth of the camera pose; 
the \emph{Aroa's room} dataset, a child's real room with dozens of different objects;
the \emph{Snack} dataset, a sequence that shows several instances of the same
object models and force camera relocation; and
the \emph{Snack with clutter} dataset, a small area with repeated objects in a small
space with occlusion and background clutter.


\subsection{Desktop dataset}
\label{sec:EvaluationDesktop}

\figTestingObjects
\figNisterObjects

The desktop dataset is a 6'26" sequence of $640\times480$ images collected
with an Unibrain camera on a desktop area, 
which we used to  test our object recognition algorithm. 
The dataset shows the 6 objects illustrated in
Figure~\ref{fig:TestingObjects}, whose largest dimension is between 10 and 20 cm.   
These were modeled with consumer photo cameras.
In addition to these objects, we created models from the
image dataset provided by Nister \& Stewenius \cite{Nister06}. These
are sets of 4 images depicting general objects, as those shown
in Figure \ref{fig:NisterObjects}, under different points of view and illumination
conditions. We used up to 494 sets of images to populate our
object databases with models to be used as distractors
for the object candidate detection step.

\figQueryScores

We show first the results of the object candidate retrieval step
from single images. 

In addition to the KL divergence, we evaluated the performance to detect
object models of other 
similarity metrics popular in bag-of-words approaches \cite{Sivic09}. 
We selected a set of 300 $640\times480$ images 
that show
one object at a time from a distance of between 20 and 70 cm.
Then, we manually masked out the background and query the database
varying the amount of stored objects, computing the KL divergence, the 
Bhattacharyya coefficient, the $\chi^2$ distance and the 
$L_1$-norm and $L_2$-norm distances. 
Figure \ref{fig:QueryScores} shows the retrieval performance of each metric,
defined as the percentage of correct object candidates returned. 
Figure \ref{fig:QueryScores:Acc} shows the performance against a database
of 500 models regarding the number of top results that we consider candidates.
The KL divergence offers a higher performance in comparison with the rest of
the metrics. We can see that the performance increases remarkably when we consider
as candidates up to the top-10 results, where the performance stalls. 
Since increasing the number of candidates will hit the execution time in the
verification step, we choose to select the top-10 object candidates for each
region of interest.
Figure~\ref{fig:QueryScores:Nobjs} shows the evolution of the metric scores
when the top-10 candidates are selected from databases comprising from 10 to 500
objects. We can see that the KL divergence always provides the best performance.

\figExperimentDesktopDetections
\figExperimentDesktopMap
\figDetectionError

When running our complete Object SLAM  approach in the desktop dataset,
the 6 objects were correctly located in the space, 
with no false positive detections.
Figure \ref{fig:ExperimentDesktopDetections} shows some correct 
object detections in single images. Figure \ref{fig:ExperimentDesktopMap}
shows the obtained map, including objects, keyframes (gray cameras),
semantic keyframes (red cameras) and points. 

In some cases, the pose of an object obtained from a single-image detection
may be inaccurate. Any algorithm is subject to this due to several factors, 
such as perceptual aliasing, or a bad geometrical conditioning. Since we do not 
rely on a single detection to locate an object in the space, our system is
able to overcome from detection inaccuracies. For example, 
consider the case shown by Figure~\ref{fig:DetectionErrorA}. 
We have two prior locations of the chewing gum box and a card (blue outline),
from which the two poses of the objects are obtained (red outline). 
However, the points observed of the card (red dots) are not widely distributed,
causing the recovered pose to be ill conditioned. 
This results in an inaccurate object pose although the 2D-to-3D correspondences
are correct. Since we had additional geometrical information accumulated
of previous observations of the card, its actually computed pose remained correct,
as can be seen in the next detection, shown by Figure~\ref{fig:DetectionErrorB},
where the recognition of the card is fully accurate.

\figExecutionTime


\figRGBDSLAM

The sequence is processed in real time. Figure~\ref{fig:ExecutionTimeTracking} shows
the execution time of the SLAM tracking (block averaged for readability), 
which takes $3.3$ ms on average. Figure~\ref{fig:ExecutionTimeRecognition} shows
the execution time taken by the object recognition process with each image,
being $138$ ms per image on average. Since the tracking and the recognition
run in parallel, the SLAM map is created successfully in real time 
independently of the time taken by the object recognition.
It is worth mentioning that the total execution time of our system, which
performs object recognition and SLAM, is lower to the time consumption of
other approaches that run object recognition only, as we show
in Table~\ref{tab:ExecutionTime}.  
MOPED \cite{Collet2011} is a state-of-the-art algorithm,
highly optimized to use thread parallelization, 
that recognizes 3D objects from SIFT features and retrieve its pose in the space
from single images. To obtain its results, we ran MOPED in the desktop dataset, 
on the same computer and with the same 500 models. 
Our object recognition algorithm yields similar results to those by MOPED,
as shown in Table~\ref{tab:PriorsMoped} (\emph{our system, no priors}).
Furthermore, this table shows that our full system is able to provide 
more detections of objects when we exploit the prior locations obtained by
the SLAM optimization over time.
The advantage of the prior information is well illustrated with the lion toy.
It is a challenging object to recognize because
its repetitive red and white stripes are prone to cause mismatches. Our 
algorithm triangulated its initial pose from 4 observations, setting a prior
location that enabled subsequent successful detections.
This highlights the fact that any object recognition algorithm can be enhanced
by the SLAM approach we propose.

\tabExecutionTime
\tabPriorsMoped

\figGTmap


\subsection{RGB-D SLAM Dataset}

Sturm\etal\cite{sturm12iros} acquired several video sequences with a RGB-D Kinect camera to evaluate RGB-D SLAM systems. These are conveniently provided with the ground truth trajectory of the camera, obtained from a high-accuracy motion-capture system. We utilized this dataset to measure the accuracy on camera location by our system with monocular images. 

We made use of several of their sequences: one to evaluate our Object SLAM, and the others to train the models of the objects that appear in the former. We ran our system on the sequence titled \emph{freiburg3 nostructure texture near withloop} in which the camera describes a loop, moving around some posters lying on the floor (shown by Figure~\ref{fig:RGBDSLAM}). We built the object models with the \emph{validation} version of the previous sequence and with the sequence \emph{freiburg3 nostructure texture far}, which shows the same posters but from different camera positions and distances. 
Thus, we do not use the same data for training and evaluation.
We created the object models by taking sparse RGB images of each poster ($\sim$20) and processing them as explained in Section~\ref{sec:ObjectModels}. We set their scale by reconstructing their 3D point clouds and measuring the real distance between pairs of points. As in the desktop dataset, besides these 8 models, we filled the database up to 500 models.

\tabGTerror
\figGTtrajectory 

Our system is able to recognize and place all the posters in the scene but two: the smallest one, for which no detections are obtained due to its size, and the one in the middle of the scene, because the camera moves very close to the floor and it barely focuses the center of the trajectory. 

The produced map is shown in Figure~\ref{fig:GTmap}. Its scale is successfully estimated from the triangulations of the objects. The average error obtained in the poses of the keyframes (the only poses that
are optimized by the BA) is $3.4$ cm in translation and $1.4$ degrees in rotation.

\tabRPE
\figExecutionTimeRgbd

We compared our system with the RGB-D SLAM algorithm \cite{endres2012evaluation} and PTAM \cite{Klein07}. 
RGB-D SLAM creates a graph with the poses of the camera, linking the nodes with the relative transformation between them. These are obtained by computing 3D-to-3D correspondences from SURF features between pairs of images, by using the RGB and depth data of a Kinect camera. The graph is then optimized with g2o \cite{Kummerle11}. Since their sensor provides depth, their map is at real scale. 

A qualitative comparison of the trajectories estimated by our system, RGB-D SLAM and PTAM is shown in Figure~\ref{fig:GTtrajectory} along with the ground truth. We obtained this figure by aligning, according to the timestamps of the images, the evaluated trajectories with the ground truth by means of Horn's method \cite{Horn87}. Since PTAM is a monocular system, we computed the scale by aligning the first meter of the trajectory in 7DoF.

For a quantitative comparison we ran the sequence on each system 10 times and report the average Absolute Trajectory Error (ATE) and Relative Pose Error (RPE) \cite{sturm12iros}. ATE compares the absolute distances between the estimated and the ground truth trajectories after alignment; the results are on Table~\ref{tab:GTerror}. For the rotation error, we computed the circular mean and standard deviation of the angles between the orientation of each pose with its corresponding one in the ground truth. We also show the root-mean-square error (RMSE) on translation. We report the relative pose error on Table~\ref{tab:RPE}, which measures the local accuracy of the trajectory over a fixed time interval and corresponds to the drift of the trajectory. Instead of restricting to evaluate in a fixed time interval, we compute the average over all possible time intervals \cite{sturm12iros}. 
The average ATE yielded by RGB-D SLAM is $9.6$ cm in translation and $3.9$ degrees in rotation. We observed this system creates a bias in the scale of the trajectories in some datasets that is producing this error. However, the origin of this bias is not clear.
We conclude that by introducing objects, our monocular system can retrieve the real scale of the scene and create maps that are more accurate and contain richer information (3D objects and points).


Figure~\ref{fig:ExecutionTimeTrackingRgbd} shows the execution time of SLAM tracking, which takes $7.6$ ms on average. The execution time taken by the object recognition process is $220$ ms per image on average with a database of 500 objects, and it is shown in Figure~\ref{fig:ExecutionTimeRecognitionRgbd}.

\subsection{Aroa's Room and Snack datasets}

We collected three more sequences to show qualitative results of our system in challenging
scenarios. 

\figAroasRoom
\figAroasRoomMap

The \emph{Aroa's room} dataset was collected with a Kinect camera 
(using the RGB sensor only) in a child's real room, where
we modeled a set of 13 objects (toys and pieces of furniture, such as blinds and a wall poster) 
of diverse size, with a consumer photo camera.
Figure~\ref{fig:AroasRoom} shows the environment and some of the objects.
The main challenge of this scenario is the highly textured clutter that can produce 
mismatches in the object recognition.
Figure~\ref{fig:AroasRoomMap} shows the resulting map, where all the objects
are located. The full execution
can be watched on video\footnote{\url{http://youtu.be/cR_tkKpDZuo}}.

\figSnack
\figSnackPTAM
\figSnackMap

\figSnackClutter
\figSnackClutterOk

The \emph{Snack} dataset shows a sequence recorded with a Unibrain camera,
where 10 bottles and cans, some of them identical, are placed together on a table. 
The database is filled with 21 models
of snacks. Figure~\ref{fig:Snack} shows the 5 models that actually appear on the table.
In this sequence, we intentionally made SLAM lose tracking of the camera on two occasions. 
The observations of the objects are still merged once the camera is relocated,
obtaining successful triangulations.
Figure~\ref{fig:SnackPTAM} shows the system running. The two windows on top
show the PTAM tracking and the 3D map created so far. Below, on the left, 
the current object detections (red outline) and the prior knowledge about their
position (blue outline). Figure~\ref{fig:SnackMap} shows the objects in the
final map.
The full sequence can be watched on video\footnote{\url{http://youtu.be/C3z62h6NPt4}}.

\figSnackClutterMap
\figSnackClutterBad

The \emph{Snack with clutter} dataset shows a sequence with 6 bottles and cans, 
some of them identical, in the highly cluttered and textured scenario depicted by
Figure~\ref{fig:SnackClutter}. The main challenge of this scenario for the object recognition
is that objects are placed very close each other and with remarkable occlusion, 
so that regions of features may not separate objects accurately. 
In spite of that, the object detector yields successful results in single frames, as 
those depicted by Figure~\ref{fig:SnackClutterOk}. 
In Figures~\ref{fig:SnackClutterOk:Bio} and \ref{fig:SnackClutterOk:Pringles}
both the bottle and the can are found before any prior information is available, 
and even if only a small part of them is visible. 
Although all the objects present in those two frames are not recognized, those detections
create prior information that is exploited in next frames, making it possible to 
detect all the object, as shown by Figure~\ref{fig:SnackClutterOk:All}. 
This exhibits the ability of our system to exploit the information provided by a sequence of images 
instead of working in a single image basis.
Finally, our system produces the map shown by Figure~\ref{fig:SnackClutterMap}.
The full sequence can be watched on video\footnote{\url{http://youtu.be/u8gvKahWt1Q}}.

These three sequences show that our system can create consistent 3D maps of points and objects
handling very different objects at the same time, 
dealing with several instances of the same models, in highly cluttered and occluded scenes
and even in cases in which track is lost.

Our system provides safety checks at different stages to keep the map consistent. 
The erroneous observations are due to spurious detections. These rarely occur because of the
feature match constraints imposed in the object recognition stage. 
If they happen or the computed pose is little accurate, 
the observation accumulation stage prevents the wrong detections from damaging the map 
because several consistent observations with wide parallax are very unlikely.
Figure~\ref{fig:SnackClutterBad} illustrates this case. The blue lines outline the pose
provided by the prior information. There is a prior around each object and two additional
wrong ones around the blue bottles. These were created by inaccurate object detections.
Since these observations did not match the correct detections, 
they were accumulated as new instances of the bottle. However, they are not triangulated  
because they are not supported by other observations.
In the rare case that a wrong instance was triangulated and anchor points created, 
the Huber robust influence function (equation~\ref{eq:huber}) would decrease its
impact in the optimization stage when there were enough correct anchor points in the map, 
keeping the map consistent.

\section{Conclusions}
\label{sec:Conclusions}

We have presented an object-aware monocular SLAM system that includes a novel and efficient 3D object recognition algorithm for a database up to 500 3D object models. On the one hand, we have shown how embedding the single frame bag-of-words recognition method in the SLAM pipeline can boost the recognition performance in datasets with dozens of different objects, repeated instances, occlusion and clutter. We believe that this benefit is not only achievable by this technique but by any other recognition method embedded within the SLAM pipeline that can exploit the accumulated observations of objects.

On the other hand, inclusion of objects adds to the SLAM map a collection of anchor points that provides geometrical constraints in the back-end optimization and enables the real map scale estimation. We have shown our system can yield more accurate maps than other state-of-the-art algorithms that use RGB-D data.

There is a case we have not addressed in this work: when the first object inserted in the map is originated by wrong observations. This would cause a first incorrect scale estimate and lead to a missized map. This may be tackled by inspecting the variance of the scale estimates given by each object triangulation, so that any observation with an inconsistent scale could be eliminated. Alternatively, the problem might also be avoided if an initial rough scale estimate is available; for example, from the odometry or IMU sensors with which robots and mobile devices are usually equipped. Nevertheless, because of the safety steps of our approach, this case can rarely occur and it did not happen in our experiments. 

Including objects in maps paves the way to augment them with semantic data, providing enriched information to a user, or additional knowledge about an environment to an operating robot \cite{RoboEarthRAM11}. We can use this knowledge in a future work to reason about the mobility of objects, making it possible to allow object frames to move in the 3D space, creating dynamic maps.
\appendix

\section{Efficient KL-divergence computation}
\label{apx:kl}

\newcommand{\VV}{\mathcal{V}}
\newcommand{\VVno}{\overline{\mathcal{V}}}
\newcommand{\WW}{\mathcal{W}}
\newcommand{\WWno}{\overline{\mathcal{W}}}

Let $\vv$ and $\ww$ be two vectors such that $\lVert \vv \rVert_1 = \lVert \ww \rVert_1 = 1$,
$\mathcal{V} = \{ i \mid v_i \neq 0 \}$,  $\VVno = \{ i \mid v_i = 0 \}$, and analogously for 
$\WW$ and $\WWno$.
The Kullback-Leibler divergence is defined as
\begin{equation}
  \label{eq:kl}
  \kl(\vv, \ww) = \sum_{\VV} v_i \log \frac{v_i}{w_i}.
\end{equation}
To avoid undetermined values, we substitute $w_i$ by a constant value 
$\varepsilon ~ \rightarrow ~ 0^+$ when
$w_i = 0$, so we can rewrite \eqref{eq:kl} as
\begin{align}
  \kl(\vv, \ww) &= \sum_{\VV \cap \WW} v_i \log \frac{v_i}{w_i} +
    \sum_{\VV \cap \WWno} v_i \log \frac{v_i}{\varepsilon} \\
  &= \sum_{\VV \cap \WW} v_i \log \frac{v_i}{w_i} +
    \sum_{\VV} v_i \log \frac{v_i}{\varepsilon} -
    \sum_{\VV \cap \WW} v_i \log \frac{v_i}{\varepsilon} \\
  &= \sum_{\VV} v_i \log \frac{v_i}{\varepsilon} 
    + \sum_{\VV \cap \WW} v_i \log \frac{\varepsilon}{w_i}.
\end{align}
Since one of the addends depends only on vector $\vv$, we remove it when we
want to compare the divergence between $\vv$ (a query vector) with other 
vectors $\ww$
(object models). Therefore, our score results in
\begin{equation}
  s_{\kl}(\vv, \ww) = \sum_{\VV \cap \WW} v_i \log \frac{\varepsilon}{w_i}.
\end{equation}


\section*{Acknowledgments}
This research has been partly funded by
the European Union under project RoboEarth FP7-ICT-248942,
the DGA-FSE (T04 group), 
the Direcci\'on General de
Investigaci\'on of Spain under projects DPI2012-36070, DPI2012-32168
and the Ministerio de Educaci\'on (scholarship FPU-AP2010-2906).

Special thanks to Aroa G\'alvez for allowing us to use her room and
play with her toys.


\section*{References} 
\bibliographystyle{elsarticle-num} 
\bibliography{references}




\end{document}